\documentclass[11pt]{article}

\usepackage{acl}

\usepackage{times}
\usepackage{latexsym}

\usepackage[T1]{fontenc}
\usepackage[utf8]{inputenc}

\usepackage{microtype}

\usepackage{inconsolata}

\usepackage{graphicx}

\usepackage{amsmath,amsfonts,bm}

\def\eqref#1{equation~\ref{#1}}
\def\1{\bm{1}}

\DeclareMathAlphabet{\mathsfit}{\encodingdefault}{\sfdefault}{m}{sl}
\SetMathAlphabet{\mathsfit}{bold}{\encodingdefault}{\sfdefault}{bx}{n}

\usepackage{hyperref}
\usepackage{url}
\usepackage[skip=6pt]{caption} 
\usepackage{algorithm}
\usepackage{algorithmic}
\usepackage{amsmath}
\usepackage{newfloat}
\usepackage{listings}
\usepackage{enumitem}

\usepackage{graphicx}
\usepackage{booktabs}
\usepackage{multirow}
\usepackage{xcolor}

\usepackage{color, colortbl}

\definecolor{mygray}{gray}{0.95}

\usepackage{array}
\usepackage{tabularx}

\usepackage{tcolorbox}
\usepackage{seqsplit}
\usepackage{makecell}
\usepackage{tcolorbox}
\usepackage{textcomp}
\tcbuselibrary{theorems,breakable,skins}
\newtcbtheorem[]{myexample}{Example}%
{
  breakable,
  enhanced,
  colback=green!5,
  colframe=green!35!black,
  fonttitle=\bfseries,
  top=2pt,         % Space between top frame and content
  bottom=2pt,      % Space between bottom frame and content
  before skip=4pt, % Space before the box
  after skip=4pt   % Space after the box
}{th}
\newcommand{\ie}{\emph{i.e., }}
\newcommand{\eg}{\emph{e.g., }}

\usepackage{etoolbox}
\makeatletter
\patchcmd{\@maketitle}{\bfseries\@author}{\normalfont\@author}{}{}
\patchcmd{\@maketitle}{\large\bfseries\@author}{\large\normalfont\@author}{}{}
\patchcmd{\@maketitle}{{\bfseries\@author}}{{\normalfont\@author}}{}{}
\patchcmd{\@maketitle}{{\large\bfseries\@author}}{{\large\normalfont\@author}}{}{}
\makeatother

\title{MolReasoner: Toward Effective and Interpretable Reasoning for Molecular LLMs}
\newcommand{\equalcontrib}{\thanks{These authors contributed equally.}}
\newcommand{\corresponding}{\thanks{Corresponding author.}}

\author{
\normalfont\mdseries\upshape\selectfont
Guojiang Zhao$^{1}$\equalcontrib \quad
Zixiang Lu$^{1}$\footnotemark[1] \quad
Yutang Ge$^{1}$ \quad
Sihang Li$^{1}$ \quad
Zheng Cheng$^{2}$ \quad
Haitao Lin$^{1}$ \quad
Lirong Wu$^{1}$ \\
Hanchen Xia$^{3}$ \quad
Hengxing Cai$^{1}$ \quad
Wentao Guo$^{1}$ \quad
Hongshuai Wang$^{1}$ \quad
Mingjun Xu$^{1}$ \quad
Siyu Zhu$^{4}$ \\
Guolin Ke$^{1}$ \quad
Linfeng Zhang$^{1}$ \quad
Zhifeng Gao$^{1}$\corresponding \\
\\
$^{1}$ DP Technology, Beijing, China \\
$^{2}$ AI for Science Institute, Beijing, China \\
$^{3}$ Shanghai Jiao Tong University, Shanghai, China \\
$^{4}$ Fudan University, Shanghai, China
}

\begin{document}
\maketitle

\begin{abstract}
Large Language Models (LLMs) have shown impressive performance across various domains, but their ability to perform molecular reasoning remains underexplored. Existing methods mostly rely on general-purpose prompting, which lacks domain-specific molecular semantics, or fine-tuning, which faces challenges in interpretability and reasoning depth, often leading to structural and textual hallucinations. To address these issues, we introduce \textbf{MolReasoner}, a two-stage framework that transitions LLMs from memorization to high-fidelity chemical reasoning. In the Mol-SFT stage, knowledge-enhanced Chain-of-Thought (CoT) data provides a strong foundation, while the Mol-RL stage refines reasoning using a novel, task-adaptive reward system  to mitigate hallucinations. Extensive evaluations demonstrate that MolReasoner significantly outperforms a wide range of strong baselines in both molecule generation and captioning tasks. Further analyses highlight the framework's synergistic design and its ability to produce more interpretable outputs. Our work presents a principled and effective new approach for advancing high-fidelity molecular reasoning.

\end{abstract}


\section{Introduction}
Given the significance of molecular science~\citep{molecular-science} --- spanning applications such as drug discovery~\citep{drug-discovery, drug1,drug2} and materials design~\citep{material-design, ms1,ms2,ms3, zhao2025virtual}, --- as well as the growing need for deeper insights into molecular structures, molecular reasoning tasks have become vital for uncovering underlying chemical relationships and intrinsic patterns in molecules, directly supporting molecular design, property prediction, and the generation of novel molecules.
In parallel, large language models (LLMs)~\citep{gpt4o,gemini,qwen3} have recently achieved impressive capabilities across diverse domains, prompting interest in utilizing LLMs to enhance molecular comprehension and generation. To bridge molecular science and language modeling, a notable research direction is the translation between molecular descriptors and natural language \citep{molt5}.

\begin{figure*}[t]
    \centering
    \vspace{-0.2in}
\includegraphics[width=\textwidth]{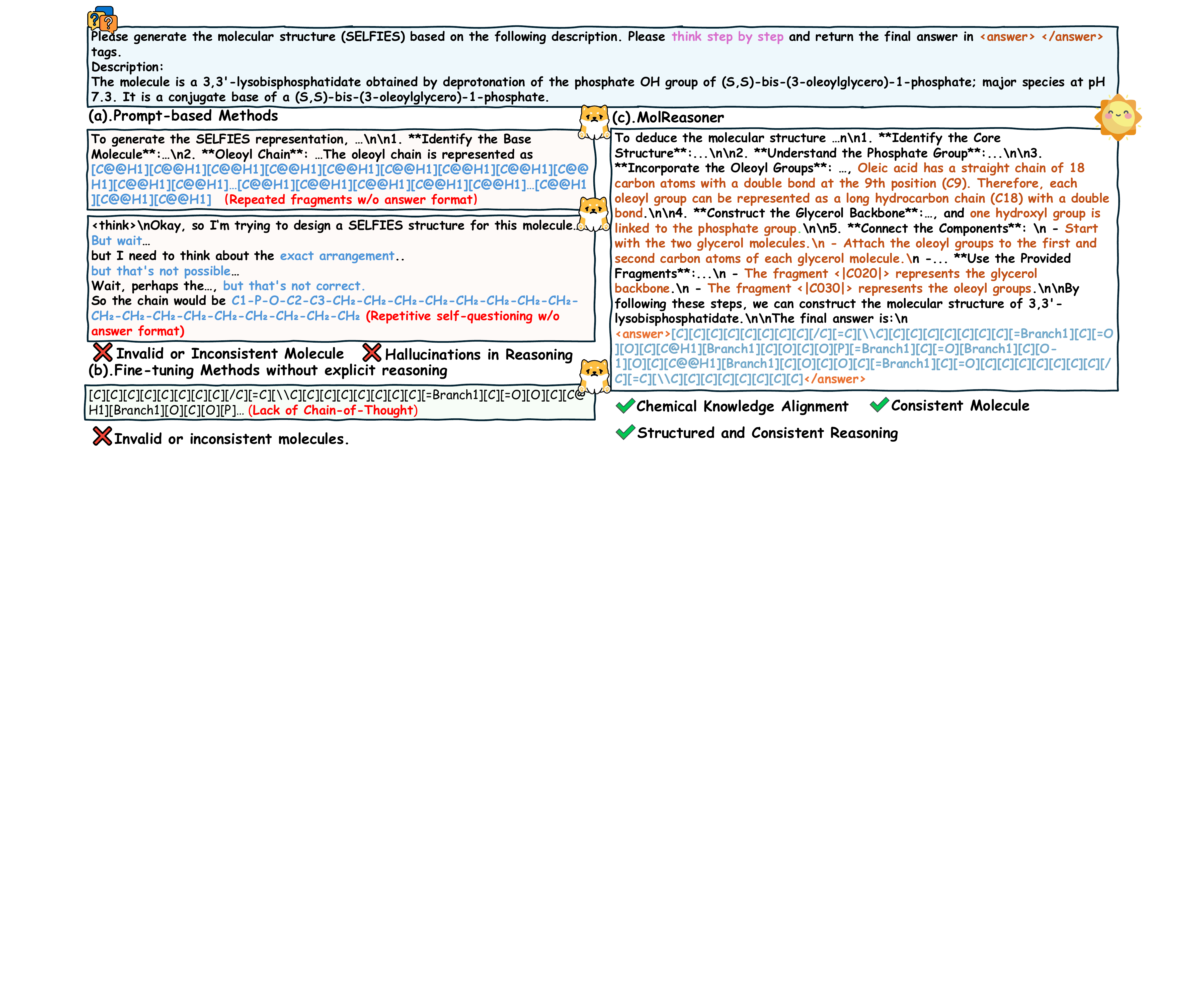}
    \captionsetup{belowskip=-10pt} % 仅对这一张图有效
    \caption{Examples of text-based molecule generation. (a) \textbf{Prompt-based methods} often hallucinate and yield chemically invalid molecules due to a lack of chemistry-specific adaptation. (b) \textbf{Fine-tuning without explicit reasoning} encourages memorization over generalization, reducing interpretability. (c) \textbf{MolReasoner} provides structure-grounded Chain-of-Thought reasoning, yielding interpretable and chemically valid candidates.}
    % \vspace{-0.1in}
    \label{fig:teaser_image}
\end{figure*}

The extensive knowledge and reasoning capabilities of LLMs make them promising candidates for complex text-related molecular tasks.
Early \textbf{Prompt-based approaches}~\citep{prompt1,prompt2,prompt3} utilize general-purpose LLMs without domain-specific fine-tuning, relying on intentionally engineered instructions or in-context examples. 
However, as illustrated in Figure~\ref{fig:teaser_image} and Appendix \ref{appx: cases}, these methods are limited in capturing molecular semantics due to the absence of chemistry-specific adaptation, leading to a heavy reliance on superficial linguistic cues or templates rather than accurately encoding underlying chemical structures. 
This issue can result in detrimental errors, including incorrect atom counts and the generation of chemically implausible structures, underscoring the limitations of general-purpose LLMs in effectively reasoning about molecular content.

Subsequently, \textbf{Fine-tuning methods without explicit reasoning}~\citep{mol-instruction,llasmol,molca,3d-molm,Mol-LLaMA} emerged, training LLMs directly on molecule-text pairs formatted as question-answer or instruction-following tasks.
However, as shown in Figure~\ref{fig:teaser_image}, these methods utilize supervision signals limited exclusively to final outputs, lacking explicit guidance through intermediate reasoning steps, causing supervised memorization rather than genuine molecular reasoning.
Therefore, the models are incapable of genuinely internalizing chemical principles and logical reasoning processes, limiting their generalization ability toward unseen molecular structures~\citep{sft-rl}. 
Furthermore, the absence of explicit reasoning undermines interpretability, which in turn reduces model reliability and restricts its applicability in high-stakes molecular applications. These limitations underscore that existing methods fail to exploit the potential of LLMs in molecular tasks, largely due to insufficient reasoning capabilities. 
Motivated by these observations, we pose and answer the central research question: \textit{Can we go beyond mere memorization and teach LLMs to reason on molecular tasks?}

A natural approach to addressing this issue is to construct Chain-of-Thought (CoT) data~\citep{cot}, which provides explicit reasoning processes alongside question-answer pairs. 
However, manually crafting such detailed reasoning is prohibitively costly, as it demands substantial input from domain experts.
Recent advances in Large Reasoning Models (LRMs)~\citep{o1,deepseek, ether0} propose an alternative by leveraging Reinforcement Learning (RL) to promote reasoning based solely on outcome-based supervision.
While this reduces the reliance on handcrafted reasoning data, it encounters another obstacle --- the \textit{cold-start} problem. 
Without prior reasoning guidance or domain-specific adaptation, models initially struggle to discover effective reasoning strategies, resulting in sparse reward signals early in training. Also, relying on outcome-based rewards in LRMs can lead to structural hallucinations, especially in molecular reasoning, where domain-specific patterns are essential. Without sufficient domain knowledge or reasoning guidance, these methods struggle to produce accurate, high-fidelity results.

To systematically address these challenges, we propose \textbf{MolReasoner}, an innovative two-stage training framework. Our core contribution is embodied in a closed-loop, synergistic paradigm that transitions from "knowledge guidance" to "multi-dimensional calibration". First, in the \textbf{Mol-SFT} stage, we leverage a knowledge-guided CoT generation process that injects structured chemical knowledge into the reasoning process, providing a high-quality reasoning starting point for the model. Second, in the \textbf{Mol-RL} stage, we design a task-adaptive, multi-dimensional reward mechanism to precisely calibrate the output. 
Comprehensive experimental evaluations not only highlight the strong performance of MolReasoner—which achieves significant gains over a wide range of strong baselines, including large-scale models, in both reasoning and generation accuracy—but also illuminate the underlying mechanisms of its success. Our extensive ablation studies point to the framework's synergistic value: the Mol-SFT stage provides a valuable starting point for reinforcement learning, while Mol-RL calibration helps the model develop a key ability to self-correct beyond imitation. These studies also suggest that our composite reward mechanisms help mitigate "reward hacking" while capturing molecular structure at different granularities, and that structured CoT facilitates more efficient learning. Furthermore, in-depth error typology analysis indicates that, in contrast to the severe hallucinations and inconsistent reasoning of baseline LLMs, MolReasoner tends to exhibit a more interpretable failure mode characterized by localized, diagnosable errors. Finally, generalization tests on out-of-distribution (OOD) datasets suggest that MolReasoner acquires a transferable, fundamental chemical reasoning capability, as its performance advantage is largely maintained on unseen data. Collectively, these findings offer strong evidence for the novelty and effectiveness of MolReasoner, presenting it as a new approach for the field of molecular captioning and generation.

% For molecule generation, we introduce our novel "fragment similarity" and "functional group matching" rewards to curb structural hallucination. For molecule captioning, we employ a composite reward that integrates multiple linguistic metrics to prevent "reward hacking" and suppress textual hallucination.

% \input{chapters/2_related_work}
% \input{chapters/3_prelim}
\section{Methodology}

\begin{figure*}[t]
    \centering
    % \vspace{-0.2in}
    \includegraphics[width=0.9\textwidth]{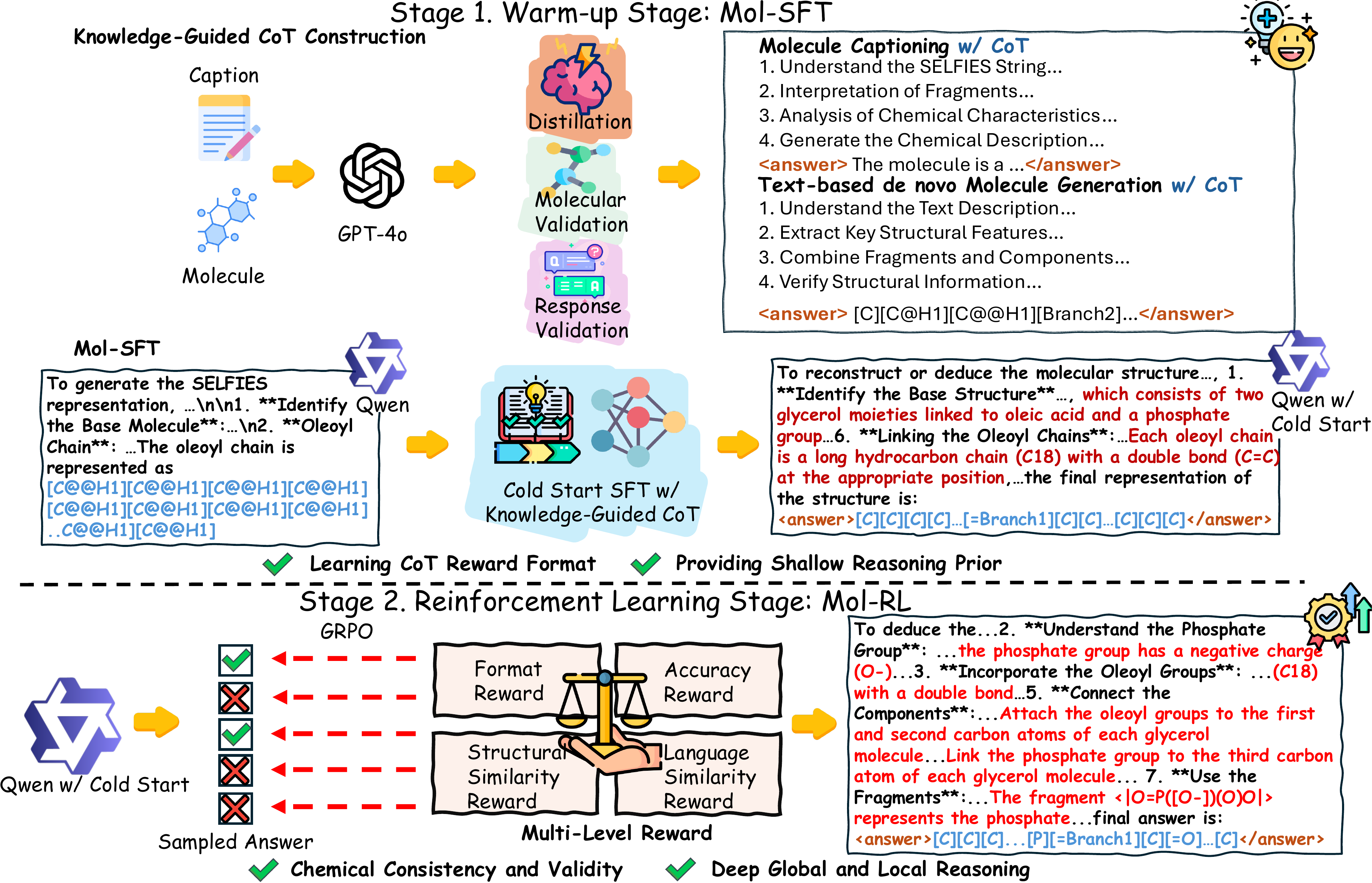}
    \caption{MolReasoner is a two-stage training framework: (1) Mol-SFT initially utilizes molecule–text pairs, augmented by reasoning trajectories generated via GPT-4o, to bootstrap reasoning capabilities; and (2) Mol-RL subsequently refines the reasoning ability through a carefully designed reward function that encourages precise alignment between molecular structures and their corresponding textual descriptions.}
    % \vspace{-0.2in}
    \label{fig:example_image}
\end{figure*}

% In the course of this work, we dedicate to the molecule-text translation task~\citep{molt5}, which evaluates a model's capability to align molecular representations with corresponding natural language descriptions through effective reasoning. Further details can be found in the appendix. Additionally, the appendix includes a comprehensive review of related research on molecular language models and large-scale reasoning models. In the following section, we present the training pipeline of MolReasoner. 
% Started with detailing the construction of CoT data for cold-start, we then introduce how the Mol-SFT and Mol-RL stages enhance the molecular reasoning abilities of the model.

In this work, we focus on the molecule-text translation task~\citep{molt5}, which evaluates a model's ability to align molecular representations with natural language descriptions through effective reasoning. Details are provided in the Appendix \ref{appx: relatedwork} and \ref{appx: task_intro}. Next, we present the MolReasoner training pipeline, starting with the construction of CoT data for cold-start and explaining how the Mol-SFT and Mol-RL stages enhance molecular reasoning.

\subsection{Knowledge-Guided CoT Data Construction}
To warm up before RL, we propose a knowledge-guided CoT data construction pipeline tailored specifically for molecular tasks, which aims to familiarize LLMs with interpretable reasoning processes. There are two molecular translation tasks: 1) molecule captioning and 2) text-based de novo molecule generation. 
Leveraging the ChEBI‑20~\citep{chembl} training set of molecule-text pairs, we construct CoT datasets using tailored prompt template, chemical knowledge injection, and rigorous output filtering strategies.

% Inspired by prior work ~\citep{jang-etal-2025-structural} incorporating structural knowledge to promote model reasoning, we further enrich our prompts with comprehensive and detailed chemical structure information.
% As showcased in the appendix, we extract statistical structural features (\eg, number of rings, aromaticity, molecular weight) to anchor the model’s understanding of molecular structures. 
% Additionally, considering the crucial role of functional groups in defining molecular properties, we adopt the EFGs~\citep{efgs} approach to identify key functional groups and hydrocarbon fragments, integrating these into prompts to strengthen the alignment between chemical structures and linguistic descriptions. 
% This multi-level guidance --- combining structural descriptors and functional fragments --- enables the model to produce coherent and chemically meaningful reasoning trajectories.

Inspired by prior work~\citep{jang-etal-2025-structural}, we enrich our prompts with detailed chemical structure information, including statistical features (e.g., rings, aromaticity, molecular weight) to anchor the model’s understanding as showcased in the Appendix \ref{appx: gpt_cot}. We also incorporate the EFGs~\citep{efgs} approach to identify key functional groups and hydrocarbon fragments, strengthening the alignment between molecular structures and linguistic descriptions. This multi-level guidance enables the model to generate coherent and chemically meaningful reasoning. While previous studies~\citep{galactica, molt5} use SMILES for molecule representation, its grammar and token order sensitivity often lead to invalid outputs. To address this, Mol-Instructions~\citep{mol-instruction} adopts the SELFIES~\citep{selfies} representation, ensuring chemical validity by eliminating structural errors such as mismatched parentheses and illogical branching.
For the molecule captioning task, we use a unified prompt template to guide GPT-4o~\citep{gpt4o} in step-by-step reasoning with the provided SELFIES sequence, generating semantically accurate captions. After strict semantic consistency filtering, we obtain around 18,000 high-quality CoT samples. For the more challenging text-based de novo molecule generation task, prompts simulate chemists' reasoning to assemble molecular components into valid SELFIES. After filtering for structural validity and semantic coherence, we acquire approximately 24,000 high-quality CoT samples. Despite both tasks using the same dataset, differences in task direction, prompt design, and filtering criteria lead to varying numbers of high-quality samples.

\subsection{Warm-up Stage: Mol-SFT}
% The effectiveness of LLMs in molecular tasks is often constrained by poor structure–semantic alignment, incoherent reasoning trajectories, and limited comprehension of chemical semantics. 
% Thus, we introduce the Molecular Supervised Fine-Tuning (\textbf{Mol-SFT}) stage, designed to establish foundational reasoning formats and linguistic logic tailored specifically to molecular tasks.

The effectiveness of LLMs in molecular tasks is often limited by poor structure–semantic alignment and incoherent reasoning. To address this, we introduce the Molecular Supervised Fine-Tuning (\textbf{Mol-SFT}) stage, which establishes foundational reasoning formats and linguistic logic tailored to molecular tasks.
In Mol-SFT, we use approximately 42,000 high-quality CoT samples from molecule captioning and text-based generation datasets to guide step-by-step reasoning. Training follows a standard autoregressive language modeling objective:
\vspace{-0.2in}
\begin{equation}
\small
\mathcal{L}_{\text{Mol-SFT}}(\theta) = - \sum_{t=1}^{T} \log p(y_t \mid x, y_{<t}; \theta)
\label{eq:mol_sft_loss}
\vspace{-0.1in}
\end{equation}
where \( x \) denotes the instruction prompt, 
\( \{y_1, y_2, \dots, y_T\} \) represents the reasoning trajectory and final answer, 
and \( \theta \) denotes the model parameters.
This stage helps the model internalize molecular reasoning formats, domain-specific terminology, and linguistic structures through example-based learning. After Mol-SFT, the model improves in following structured reasoning instructions and generating coherent reasoning chains, with shallow reasoning priors and format awareness, laying the groundwork for reinforcement learning-based refinement.

\subsection{Reinforcement Learning Stage: Mol-RL}

SFT has limitations in open-ended tasks like molecule captioning and generation, often causing the model to overfit to a single reference answer and lack diversity. To address this, we use Group Relative Policy Optimization (GRPO)~\cite{deepseekmath}, which, in our framework, undergoes a subtle yet crucial transformation. Rather than optimizing for a "unique correct answer", GRPO focuses on enhancing the "quality" of multiple possible outputs through relative advantage-based strategy optimization. In the complex task of molecule-text translation, \textbf{Mol-RL} leverages this shift to guide the model in generating diverse and structurally accurate answers, effectively mitigating both text and structural hallucination issues.
The GRPO algorithm is detailed in the Appendix \ref{appx: grpo}. For molecule captioning, we introduce a verifiable reward function $R_{\text{cap}}$ designed to encourage correct formatting and semantically accurate natural-language descriptions:
% \vspace{-0.31in}
\begin{equation}
\small
R_{\text{cap}} = \begin{cases}
0.5 + 1.5 \times R_{\text{language}}, 
& \text{if format is correct}, \\
0.0, & \text{if format is incorrect.}
\end{cases}
\label{equ:reward-caption}
\end{equation}
The format is considered correct if the final output is enclosed within \texttt{<answer>...</answer>} tags.
The language-similarity reward $R_{\text{language}}$ is computed as the mean of six standard evaluation metrics (BLEU-2, BLEU-4, METEOR, ROUGE-1, ROUGE-2, and ROUGE-L)~\citep{papineni2002bleu, banerjee2005meteor,lin2004rouge} comparing the generated caption to the reference description. In the text-based de novo molecule generation task, we propose a verifiable reward $R_{\text{gen}}$ to improve the structural integrity and semantic alignment of generated molecules:
\begin{equation}
\small
R_{\text{gen}} = \begin{cases}
0.5 + 1.5 \times R_{\text{structural}}, & \text{if format is correct}, \\
0.0, & \text{if format is incorrect.}
\end{cases}
\label{equ:reward-molgen}
\end{equation}
where the structural-similarity reward $R_{\text{structural}}$ is computed as the mean of four key components --- fingerprint similarity, SELFIES-level language similarity, fragment similarity, and functional group matching --- to provide a holistic and chemically aware measure of structural alignment. 
\begin{equation}
\small
R_{\text{structural}} = \frac{1}{4} \left( \text{FP}_{\text{sim}} + \text{SELFIES}_{\text{sim}} + \text{FRAG}_{\text{sim}} + \text{FG}_{\text{match}} \right)
\end{equation}

We compute fingerprint similarity based on three widely used molecular fingerprints: Morgan, MACCS, and RDKit~\citep{Tanimoto,fp2}. The final score is calculated as the mean of these three, providing a balanced measure of molecular similarity:
\begin{equation}
\small
\text{FP}_{\text{sim}} = \frac{1}{3} \left( \text{Morgan} + \text{MACCS} + \text{RDKit} \right)
\end{equation}
Additionally, $\text{SELFIES}_{\text{sim}}$ is assessed by computing the character-level BLEU score between the predicted and reference SELFIES sequences. 

In addition to the two similarity-based rewards, we observe that fragment and functional group hallucinations are regularly presented during the molecular reasoning process. 
Even though the generated molecules may be chemically valid, they can exhibit significant inconsistencies with the reference structures.
To address this issue, we introduce two additional rewards: fragment similarity and functional group matching. 
We use EFGs~\citep{efgs,lin2024cbgbench} to identify differing fragments between the generated molecules and the reference molecules. 
Fragment similarity is then computed by jointly considering the Jaccard overlap and fragment-level recall, effectively capturing both coverage and precision of structural subcomponents. 
Here, \( \mathcal{F}_{\text{pred}} \) and \( \mathcal{F}_{\text{ref}} \) refer to the sets of fragments in the predicted and reference molecules, respectively.
The final fragment similarity score is given by:
\begin{equation}
\small
\text{FRAG}_{\text{sim}} = 0.5 \times \frac{|\mathcal{F}{\text{pred}} \cap \mathcal{F}{\text{ref}}|}{|\mathcal{F}{\text{pred}} \cup \mathcal{F}{\text{ref}}|} + 0.5 \times \frac{|\mathcal{F}{\text{pred}} \cap \mathcal{F}{\text{ref}}|}{|\mathcal{F}{\text{ref}}|}
\end{equation}
Additionally, the functional group matching reward computes the difference in the number of functional groups, excluding CH-only groups, between the predicted and reference molecules, using an exponential decay formulation.
\begin{equation}
\small
\text{FG}_{\text{match}} = \exp\left( -\frac{ \sum_k \left| \text{count}_{\text{pred}}(k) - \text{count}_{\text{ref}}(k)) \right| }{ \sum_k \text{count}_{\text{ref}}(k) + \varepsilon } \right)
\end{equation}
where \( \text{count}_{\text{pred}}(k) \) refers to the number of occurrences of functional group \( k \) in the predicted molecule, and \( \text{count}_{\text{ref}}(k) \) refers to that of functional group \( k \) in the reference molecule. 
We set $\varepsilon=10 ^{-5}$ to ensure numerical stability. Incorporating chemical awareness and granularity through multi-level reward feedback, the model aligns chemical knowledge from global molecular semantics to local molecular structural details, ensuring greater consistency in generated chemical structure. 
%This transition from merely generating ``valid'' structures to producing ``high-quality'' structures significantly enhances its generation capabilities and semantic alignment.
As a result, the model transitions from merely producing ``valid'' molecules to generating ``high-quality'' structures that are both chemically coherent and semantically aligned with the input, improving its generation capabilities.
\section{Experiments}
This section outlines our experimental setup and presents the results that demonstrate the effectiveness of MolReasoner.
Due to the space limits, we refer to the Appendix \ref{appx: trainingsetup} for implementation details.

\subsection{Datasets}

% With validity-based filtering criteria, 

We generate approximately 42,000 high-quality CoT samples from the training set of ChEBI-20 --- 24,000 samples for text-based de novo molecule generation and 18,000 for molecule captioning. 
These samples form for the initial warm-up training stage.
For the later reinforcement learning stage, we construct two GRPO training datasets derived from ChEBI-20.

\subsection{Evaluation and Baselines}

We utilize the test set of ChEBI-20 as our benchmark. The molecules are represented using SELFIES, following Mol-Instructions~\citep{mol-instruction}.

\textbf{Molecule Captioning.} 
Following Mol-Instructions~\citep{mol-instruction}, we adopt standard language generation metrics, including BLEU, ROUGE, and METEOR, to assess the similarity between generated molecular descriptions and ground-truth references.

\textbf{Text-based de novo Molecule Generation.} 
We use RDKit to validate the chemical correctness of generated strings and compute exact match rates. In addition, we evaluate molecular similarity using Tanimoto scores, Levenshtein distance, and BLEU scores. To further assess the structural fidelity of generated molecules, we propose three fragment-level metrics: Frag-J, Frag-R, and FG-Match (Functional Group Matching). For detailed metric definitions, see Appendix \ref{appx: metrics}. Unlike Mol-Instructions, which evaluates only valid molecules, we report all metrics across the entire set of generated molecules to capture overall model performance.

\textbf{Baselines.}
Our primary goal is to examine how general-purpose LLMs can be adapted for molecular reasoning when equipped with explicit reasoning strategies. We compare our model against a range of leading LLM-based baselines, including both prompt-based methods and fine-tuning approaches, with additional comparisons to their fine-tuned versions. We also evaluate the chemical reasoning model ether0. Detailed information about the models, including configurations and training setups, can be found in Appendix \ref{appx: trainingsetup}.

% For the baseline models,
% we deliberately select baselines that are LLM-based models.
% For prompt-based methods, we compare our model against leading general-purpose LLMs, including GPT-4o, GPT-4o-mini, Deepseek-r1-0528, Qwen2.5-7B-Instruct, DeepSeek-R1-Distill-Qwen-7B, LLaMA3.1-8B-Instruct, Qwen3-8B, Qwen2.5-32B-Instruct, LLaMA3.1-70B-Instruct, Qwen2.5-72B-Instruct, and LLaMA3 with MSR(10-shot). 
% For fine-tuning approaches without explicit reasoning, we evaluate Mol-Instructions + LLaMA2/3 (without fine-tuning), Mol-LLaMA, and their fine-tuned versions: Mol-Instructions + LLaMA3 (ft) and LLaMA3 + MSR (ft). Also, we compare  chemical reasoning model ether0. We use Qwen2.5-7B-Instruct as our base model; 

\subsection{Main Results}

% 展示
\begin{table}[t]
 \centering
\scalebox{0.50}{
 \setlength{\tabcolsep}{2.2pt}
 \begin{tabular}{@{}lccccccc@{}}
   \toprule
   \textbf{Method} & \textbf{Size} & \textbf{BL.2↑} & \textbf{BL.4↑} & \textbf{ME.↑} & \textbf{RO.1↑} & \textbf{RO.2↑} & \textbf{RO.L↑} \\ \midrule
   \multicolumn{8}{c}{\textit{Closed-Source Model}} \\ \midrule
   GPT-4o & - & 0.1198 & 0.0433 & 0.1656  & 0.2323  & 0.0735 & 0.1789  \\
   GPT-4o-mini & - & 0.1084 & 0.0401  & 0.1550  & 0.2312 & 0.0719  & 0.1776 \\
   Deepseek-R1  & - & 0.1022 & 0.0354  & 0.2189   & 0.2358 & 0.0666   & 0.1666 \\ 
\midrule
    \multicolumn{8}{c}{\textit{Open-Source Model}} \\ \midrule
    Qwen2.5-7B-Instruct & 7B & 0.0839  & 0.0287 & 0.2125 & 0.2147 & 0.0633 & 0.1530 \\
    DeepSeek-R1-Qwen-7B & 7B & 0.1177  & 0.0469  & 0.1540  & 0.2212 & 0.0747 & 0.1696  \\
     LLaMA3.1-8B-Instruct & 8B & 0.1687 & 0.0784 & 0.2180 & 0.2838 & 0.1199  & 0.2272 \\
    Qwen3-8B & 8B & 0.0972 & 0.0288  & 0.1729 & 0.2067 & 0.0502 & 0.1566 \\
    Qwen2.5-32B-Instruct & 32B & 0.0950 & 0.0288  & 0.2032 & 0.2278   & 0.0637 & 0.1614   \\
     LLaMA3.1-70B-Instruct & 70B & 0.1485 & 0.0664 & 0.1845 & 0.2753  & 0.1071 & 0.2211  \\
    Qwen2.5-72B-Instruct & 72B & 0.1512 & 0.0638 & 0.1933  & 0.2715 & 0.0936 & 0.2054 \\
    LLaMA3 + MSR 10-shot & 8B & 0.1843 & 0.1068 & 0.2374 & 0.3142 & 0.1497  & 0.2525  \\
    Mol-Instructions +  LLaMA2 w/o ft & 7B & 0.1077 & 0.0750  & 0.1900  & 0.2789 & 0.1811 & 0.2569  \\
    Mol-Instructions +  LLaMA3 w/o ft & 8B & 0.1387 & 0.0951  & 0.1801 & 0.2025 & 0.1163  & 0.1694 \\
    
    Mol-LLaMA & 8B & 0.0821  & 0.0206   & 0.1305 & 0.2316 & 0.0610  & 0.1875 \\
    ether0 & 24B & 0.0153 & 0.0120  & 0.1001 & 0.1064 & 0.0104 & 0.0162 \\
    Mol-Instructions +  LLaMA3 w/ ft & 8B & 0.2590 & 0.1995 & 0.4341  & 0.3906 & 0.2472  & 0.2306  \\

    LLaMA3 + MSR w/ ft  & 8B & \underline{0.2792}  & \underline{0.2151} & \underline{0.4701}  & \underline{0.4140} & \underline{0.2655}  & \underline{0.3428}   \\

    \rowcolor{mygray}
    \textbf{MolReasoner (Ours)} & 7B & \textbf{0.4394} & \textbf{0.3233} & \textbf{0.4767 } & \textbf{0.5536} & \textbf{0.3674} & \textbf{0.4827} \\
    \bottomrule
  \end{tabular}}
  \caption{Performance of Molecule Captioning, where "ft" denotes fine-tuning. MolReasoner outperforms all closed‑source and open‑source baselines across all metrics. BL., RO., and ME. stand for BLEU, ROUGE, and METEOR, respectively. }
  \label{tab:molecule_captioning}
  \vspace{-0.27in}
\end{table}

\begin{table*}[t]
  \centering
  \scalebox{0.50}{
  \setlength{\tabcolsep}{2.2pt}
  \begin{tabular}{@{}lccccccccccc@{}}
    \toprule
    \textbf{Method} & \textbf{Size} & \textbf{BL.↑} & \textbf{Ex.↑} & \textbf{Le.↓} & \textbf{RDK.↑} & \textbf{MA.↑} & \textbf{MO.↑} & \textbf{Frag-J↑} & \textbf{Frag-R↑} & \textbf{FG-Match↑} & \textbf{Val.↑} \\ \midrule
    \multicolumn{12}{c}{\textit{Closed-Source Models}} \\ \midrule
    GPT-4o & - & 0.1723 & 0.0062 & 50.2363  & 0.0928  & 0.2064  & 0.0844 & 0.1282 & 0.1740 & 0.3678 & 0.3224 \\ 
    GPT-4o-mini & - & 0.0532  & 0.0047 & 48.0824  & 0.0858  & 0.2057  & 0.0875 & 0.0928  & 0.1235  & 0.3860 & 0.1932  \\ 
    Deepseek-R1 & - & 0.0145 & 0.0063  & 55.3267 & 0.0408  & 0.0890  & 0.0382  & 0.0372 & 0.0455  & 0.3906  & 0.6932   \\ 
\midrule
    \multicolumn{12}{c}{\textit{Open-Source Models}} \\ \midrule
    Qwen2.5-7B-Instruct & 7B & 0.0001 & 0.0031 & 35.2666 & 0.1022 & 0.2213 & 0.0804 & 0.1052 & 0.1431 & 0.3536  & 0.1990 \\ 
    DeepSeek-R1-Qwen-7B & 7B & 0.0000 & 0.0022  & 49.8502  & 0.0732  & 0.1361  & 0.0567 & 0.1190  & 0.1502  & 0.3843  & 0.0665  \\ 
     LLaMA3.1-8B-Instruct & 8B & 0.0133 & 0.0026 & 40.8965 & 0.0580 & 0.1504  & 0.0473 & 0.0745 & 0.0964 & 0.3597  & 0.2375  \\ 
    Qwen3-8B & 8B & 0.0000 & 0.0034 & 26.1532 & 0.3578 & 0.4605 & 0.3034 & 0.3504  & 0.3728  & \underline{0.5299} & 0.0119 \\ 
    Qwen2.5-32B-Instruct & 32B & 0.0054  & 0.0042  & 33.7951 & 0.1204 & 0.2566 & 0.1163 & 0.1257   & 0.1586  & 0.3542 & 0.1722 \\ 
     LLaMA3.1-70B-Instruct & 70B & 0.0819 & 0.0042 & 44.6099 & 0.0831 & 0.2344 & 0.0775 & 0.1367 & 0.1945 & 0.3522  & 0.4665 \\ 
    Qwen2.5-72B-Instruct & 72B & 0.0000 & 0.0044 & \textbf{17.8122} & 0.1496 & 0.3385 & 0.1347  & 0.1579 & 0.2160  & 0.3352 & 0.1173  \\ 
    LLaMA3 + MSR 10-shot  & 8B & 0.3134 & 0.0019 & 43.7624  & 0.1419  & 0.3004  & 0.0966 & 0.1955 & 0.2847 & 0.2946  & 0.7889  \\ 
    Mol-Instructions+  LLaMA2 w/o ft & 7B & 0.3049 & 0.0437  & 39.4265  & 0.2914 & 0.4394 & 0.2524 & 0.3326 & 0.4124  & 0.4323 & \textbf{0.9991} \\ 
    Mol-Instructions+  LLaMA3 w/o ft & 8B & 0.3323 & 0.0738 & 38.1494 & 0.3598 & 0.4956 & 0.3167 & 0.4071 & 0.4714 & 0.5056 & 0.9707 \\ 
    
    Mol-LLaMA  & 8B & - & - & -  & -  & -  & - & - & - & -  & -  \\ 
   ether0  & 24B & - & - & -  & -  & -  & - & - & - & -  & -  \\ 
   Mol-Instructions + LLaMA3 w/ ft  & 8B & 0.3343 & 0.0740 & 37.7033  & 0.3799  & 0.5096  & 0.3246 & 0.4157 & 0.4728 & 0.5130  & 0.9714  \\ 
     LLaMA3 + MSR w/ ft  & 8B & \underline{0.3382} & \textbf{0.1073} & 31.2113  & \underline{0.4013}  & \underline{0.6518}  & \underline{0.3425} & \underline{0.4569} & \underline{0.5361} & 0.5197 & \underline{0.9801}  \\ 
    \rowcolor{mygray}
    \textbf{MolReasoner (Ours)} & 7B & \textbf{0.7832} & \underline{0.0746} & \underline{26.0237} & \textbf{0.4369} & \textbf{0.6762} & \textbf{0.3618} & \textbf{0.5221} & \textbf{0.6419} & \textbf{0.5390} & 0.9655 \\ 
    \bottomrule
    
  \end{tabular}}
  \caption{Performance of Text-based de novo Molecule Generation, where "ft" denotes fine-tuning. MolReasoner surpasses both closed‑source and open‑source baselines across nearly all metrics. BL., Ex., Le., RDK., MA., MO., and Val. stand for BLEU, Exact, Levenshtein, RDK FTS, MACCS FTS, MORGAN FTS,  and Validity, respectively.}
  \label{tab:text_molecule_generation}
  \vspace{-0.2in}
\end{table*}

In this experiment, we compare MolReasoner with prompt-based methods and fine-tuning methods without explicit
reasoning, evaluating its performance on two tasks: 1) molecule captioning (Table \ref{tab:molecule_captioning}) and 2) text-based de novo molecule generation (Table \ref{tab:text_molecule_generation}). 
To ensure the robustness of our findings, all experiments were conducted \textbf{three} times with different random seeds. For clarity and due to space constraints, the tables present the \textbf{mean} values of these runs.
Across all tables, \textbf{bold} indicates the best and \underline{underline} the second-best results.
The results demonstrate that MolReasoner shows significant advantages in both reasoning capability and generation quality in two tasks. 
In molecule captioning, MolReasoner achieves the highest scores for BLEU, METEOR, and ROUGE, indicating a significantly enhanced ability to generate accurate and semantically relevant descriptions for molecules.
This strong performance extends to the text-based de novo molecule generation task. Beyond leading in textual fidelity metrics like BLEU, MolReasoner obtains the best results on crucial chemical similarity metrics, including MACCS FTS, RDK FTS, and fragment-based scores.

Prompt-based methods without domain adaptation often struggle with semantic understanding and structural accuracy due to a lack of in-depth chemical knowledge, leading to issues like functional group hallucinations, valid but semantically incorrect SELFIES, and ignoring structural details. MolReasoner, by guiding the reasoning process, generates more accurate and chemically plausible descriptions.
Models such as DeepSeek-R1-Distill-Qwen-7B \cite{guo2025deepseek}, Qwen3-8B \cite{qwen3}, and Qwen2.5-72B-Instruct \cite{bai2025qwen2}, despite incorporating reasoning during pretraining, still suffer from conflicting reasoning chains, semantically collapsed CoT, and misleading captions, as illustrated in the Appendix \ref{appx: cases}.  Compared to fine-tuning without explicit reasoning method, MolReasoner improves the accuracy and semantic consistency of molecular descriptions by introducing reasoning-enhanced mechanisms. In molecule captioning, MolReasoner outperforms fine-tuned models, and even surpasses specialized reasoning models based on GRPO, demonstrating that in such complex tasks, simple binary rewards (correct/incorrect) cannot address the challenges of molecular reasoning. In tasks like text-based de novo molecule generation, MolReasoner strengthens the reasoning process, ensuring that the generated molecules are chemically logical and structurally consistent, thus offering better reliability.

\begin{table}[t]
  \centering
  \scalebox{0.50}{
  \setlength{\tabcolsep}{2.2pt}
    \begin{tabular}{@{}lccccccc@{}}
      \toprule
      \textbf{Method} & \textbf{Size} & \textbf{BL.2↑} & \textbf{BL.4↑} & \textbf{ME.↑} & \textbf{RO.1↑} & \textbf{RO.2↑} & \textbf{RO.L↑} \\ \midrule
      \multicolumn{7}{c}{\textit{\textbf{Closed-Source Model}}} \\ \midrule
      GPT-4o & - & 0.1198  & 0.0433 & 0.1656  & 0.2323  & 0.0735 & 0.1789  \\  \midrule
      \multicolumn{7}{c}{\textit{\textbf{Ours}}} \\ \midrule
      Warm-up & 7B & 0.3829  & 0.2695  & 0.4179 & 0.4977  & 0.3104  & 0.4312 \\
      + FAR & 7B & 0.3824 & 0.2694  & 0.4180  & 0.4974  & 0.3093 & 0.4306 \\
      + BL.2 & 7B & \underline{0.4388} & 0.3163  & 0.4731  & 0.5483  & 0.3593 & 0.4762 \\
      + BL.4 & 7B & 0.4376  & \underline{0.3206} & 0.4747 & 0.5505 & 0.3643  & 0.4798  \\
      + ME. & 7B & 0.4378 & 0.3195  & \underline{0.4751 } & 0.5458  & 0.3602 & 0.4754  \\
      + RO.1 & 7B & 0.4380  & 0.3208 & \underline{0.4751 } & 0.5520 & 0.3633 & 0.4784  \\
      + RO.2 & 7B & 0.4370 & 0.3208  & 0.4711 & \underline{0.5531} & \underline{0.3652 } & \underline{0.4803 } \\ 
      Zero-RL + $R_{\text{language}}$ & 7B & 0.1333 & 0.0579  & 0.2371  & 0.2784  & 0.1027  & 0.2081  \\
      \rowcolor{mygray}
      MolReasoner + $R_{\text{language}}$ & 7B & \textbf{0.4394 } & \textbf{0.3233} & \textbf{0.4767 } & \textbf{0.5536} & \textbf{0.3674 } & \textbf{0.4827 } \\
      \bottomrule
    \end{tabular}
    }
  \caption{Progressive reward composition ablation study of different reward functions and the effect of the warm-up stage for Molecule Captioning. "Warm-up" denotes the base model without reinforcement learning.
"FAR" is Format Accuracy Reward. 
BL., RO., and ME. stand for BLEU, ROUGE, and METEOR rewards, respectively.
"Zero-RL + $R_{\text{language}}$" trains without warm-up, while "MolReasoner + $R_{\text{language}}$" is the final model. MolReasoner achieves the best performance with $R_{\text{language}}$, highlighting its superiority in captioning.}
  \label{tab:ablations_molecule_captioning}
  \vspace{-0.2in}
\end{table}

\subsection{Multidimensional Evaluation and Qualitative Analysis}

\begin{figure}[t]
    \centering\hspace{0.4in}\includegraphics[width=\linewidth]{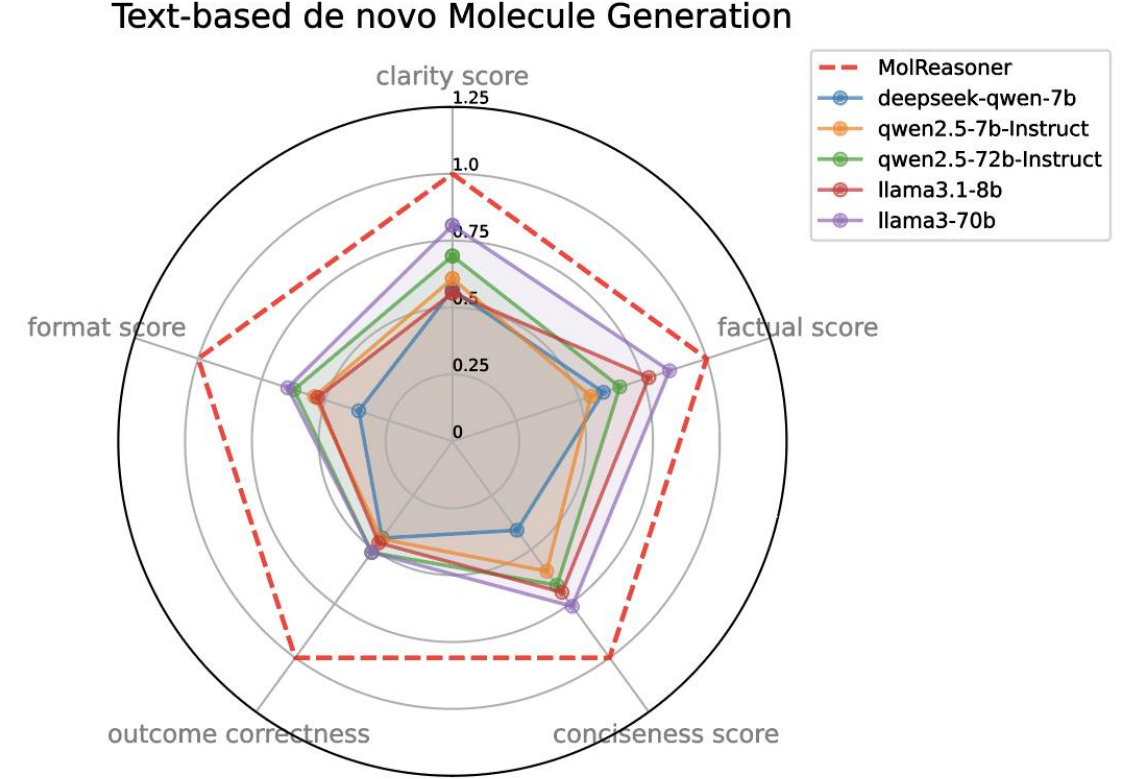} 
    \caption{\textcolor{black}
    {Performance of all models across five key evaluation metrics in the Text-based de novo Molecule Generation. To provide a more intuitive comparison, all scores are normalized by dividing them by the scores of MolReasoner.}}
    \label{fig: radar_mol1}
    % \vspace{-0.2in}
\end{figure}

To obtain a more comprehensive evaluation of model behavior, we design a multidimensional scoring scheme that assesses both reasoning and final answers. Beyond standard task-specific metrics, we consider five dimensions: \textbf{Clarity of logic}, \textbf{Factual correctness}, \textbf{Conciseness}, \textbf{Format correctness}, and \textbf{Outcome correctness}. Domain experts first manually score a subset of responses, yielding a gold-standard set that we use as few-shot exemplars in a detailed prompt (Appendix~\ref{appx: gpt_cot}) for a strong large language model, GPT-5, which we employ as an automatic judge. After this auto-scoring step, we randomly sample 30 responses for re-evaluation by experts and compute the mean absolute difference between expert and GPT-5 scores; this deviation is small (on the order of 0.38 on a 1–5 scale) and falls well within the empirical error bar of human–human variance ($\pm 0.45$), indicating that GPT-5 behaves comparably to a human rater. The resulting scores are visualized as radar plots in Figure~\ref{fig: radar_mol1} and Appendix Figure~\ref{fig: radar_mol2}, providing an intuitive comparison across the five dimensions. These plots show that MolReasoner’s advantage lies not only in answer accuracy but also in its reasoning: the generated chains-of-thought are more readable, more chemically precise, and more consistently aligned with domain knowledge than those of the baselines.

% \begin{figure*}[t]
%     \centering
%     \includegraphics[width=\textwidth]{figures/radar.pdf}
%     \vspace{-16pt}
%     \caption{Performance of all models across five key evaluation metrics in the two tasks: Text-based de novo Molecule Generation and Molecule Captioning. To provide a more intuitive comparison, all scores are normalized by dividing them by the scores of MolReasoner.}
%     \vspace{-8pt}
%     \label{fig:radar}
% \end{figure*}

\subsection{Reasoning–Answer Consistency and Error Typology}
This section provides an in-depth error analysis of MolReasoner and generic LLM baselines on molecular tasks. This error analysis reveals a key distinction: while generic LLMs often fail due to catastrophic hallucinations and incoherent reasoning, MolReasoner fails more gracefully, with errors that are typically logical and localized. This diagnosable failure pattern proves MolReasoner to be not just more performant, but also significantly more reliable and trustworthy than its black-box counterparts.
We provide details in Appendix \ref{app:reasoning_error_analysis}.

\subsection{Ablation Study}
\subsubsection{Individual Reward Ablation}

\begin{figure}[t]
    \centering\includegraphics[width=\linewidth]{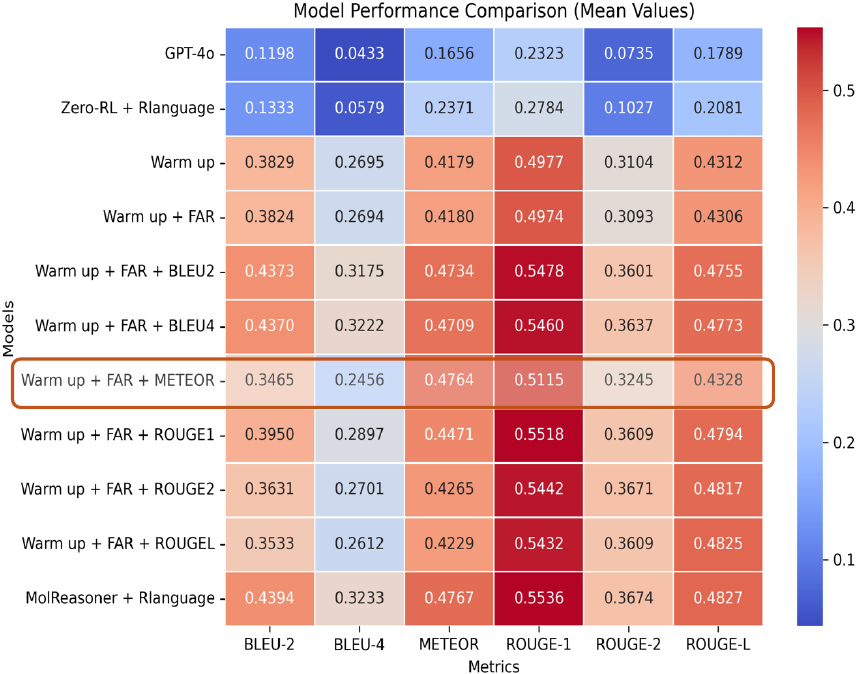}
    \caption{\textcolor{black}{Individual Reward Ablation For Molecule Captioning.}}
    \label{fig: reward1}
\end{figure}

To validate our reward design, we conducted extensive individual reward ablation studies for both tasks. For molecule captioning (Figure \ref{fig: reward1}), we observe a "reward hacking" phenomenon when optimizing for a single linguistic metric, such as METEOR. While it boosts METEOR, it significantly harms BLEU and ROUGE scores, demonstrating the bias of single-metric optimization. In contrast, our composite reward model, averaging all metrics, achieves superior and balanced performance across all criteria. For text-based de novo molecule generation (Appendix Figure \ref{fig: reward2}), the ablation study shows that our structural reward components—$\text{FP}_{\text{sim}}$, $\text{FRAG}_{\text{sim}}$, and $\text{FG}_{\text{Match}}$—complement each other by targeting specific types of structural hallucinations, providing nuanced supervision than traditional fingerprints and enhancing molecule generation accuracy.

\subsubsection{Reward Composition Ablation}

% To validate our overall training strategy, we conducted a progressive reward composition ablation for both tasks, with results detailed in Table \ref{tab:ablations_molecule_captioning} and Table \ref{tab:ablations_text_guided_molecule_generation}.  The results show the foundational importance of the warm-up stage. In both tasks, the Warm-up model demonstrates a massive performance leap over the powerful GPT-4o baseline. This confirms that our knowledge-guided CoT fine-tuning is the critical first step, equipping the model with essential domain knowledge and a strong initial policy before reinforcement learning begins. Besides, the progressive stacking of rewards demonstrates a clear synergistic effect. For molecule captioning, adding individual linguistic rewards incrementally boosts performance. Crucially, MolReasoner—employing the composite $R_{\text{language}}$ reward—achieves the highest scores across all metrics, proving that a balanced, multi-faceted reward creates a holistically superior model rather than just an expert on a single metric.
% For de novo generation, the step-by-step addition of structural rewards confirms they are complementary, not redundant. Each component progressively enhances chemical fidelity at a different granularity, validating our multi-faceted approach to ensuring structural integrity. Also, the consistently poor performance of the Zero-RL models in both tables underscores the absolute necessity of the warm-up stage. It shows that reinforcement learning is  is ineffective at learning complex, domain-specific tasks from scratch. 

To validate our training strategy, we conducted a progressive reward composition ablation for both tasks (Table \ref{tab:ablations_molecule_captioning} and Appendix Table~\ref{tab:ablations_text_guided_molecule_generation}). The results highlight the foundational importance of the warm-up stage. In both tasks, the Warm-up model significantly outperforms the powerful GPT-4o baseline, confirming that knowledge-guided CoT fine-tuning is crucial for equipping the model with domain knowledge before reinforcement learning. The progressive stacking of rewards shows a clear synergistic effect. In molecule captioning, adding linguistic rewards incrementally boosts performance, with the composite $R_{\text{language}}$ reward achieving the best results. For de novo generation, structural rewards are complementary, each enhancing chemical fidelity at different granularities. The consistently poor performance of the Zero-RL models further underscores the necessity of the warm-up stage, demonstrating that reinforcement learning alone cannot effectively tackle complex, domain-specific tasks.

\subsubsection{Impact of Molecular Representation: SMILES vs. SELFIES}
% \begin{figure}[t]
%     \vspace{-0.2in}
%     \centering\includegraphics[width=0.8\linewidth]{figures/selfies.pdf} 
%     \caption{\textcolor{black}
%     {Impact of Molecular Representation.}}
%     \label{fig: selfies}
%     \vspace{-0.1in}
% \end{figure}

To validate our choice of SELFIES, we fine-tuned two identical base models on the molecule captioning CoT dataset, differing only in the molecular representation: SMILES vs. SELFIES. As shown in Appendix Figure \ref{fig: selfies}, the SELFIES-based model outperforms the SMILES-based model across all metrics, with a more than 4.5x improvement in BLEU-4 and over a 3x increase in ROUGE-L.

\subsubsection{Impact of Knowledge-Guided CoT Data}
\begin{figure}[t]
    \centering\includegraphics[width=0.9\linewidth]{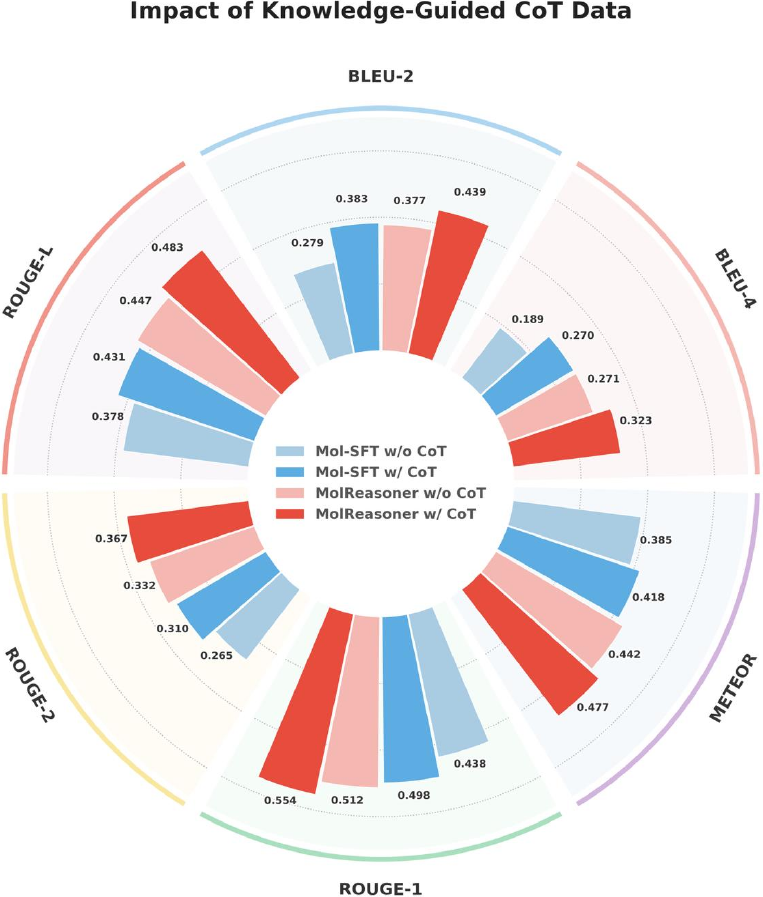} 
    \caption{\textcolor{black}
    {Impact of Knowledge-Guided CoT Data.}}
    \label{fig: cot}
    % \vspace{-0.28in}
\end{figure}

To assess the effectiveness of our knowledge-guided CoT dataset, we conducted ablation studies on the Molecule Captioning task, comparing models with and without CoT annotations: "Mol-SFT w/o CoT" vs. "Mol-SFT w/ CoT" and "MolReasoner w/o CoT" vs. "MolReasoner w/ CoT". As shown in Figure \ref{fig: cot}, CoT annotations improved performance across all key metrics, validating that our CoT trajectories enable deeper chemical reasoning and enhance molecular description accuracy. Additionally, the MolReasoner framework shows a synergistic effect with CoT annotations.

\subsection{Out-of-Distribution Evaluation}
We evaluated MolReasoner's generalization ability through Out-of-Distribution (OOD) testing on general benchmarks (HellaSwag~\cite{zellers_hellaswag:_2019}, ARC-easy~\cite{ARC}, OpenBookQA~\cite{OpenBookQA2018}, SciQ~\cite{SciQ}) and the molecular OOD dataset TOMG. For general datasets, we used 5-shot reasoning, and for TOMG~\cite{li2024tomg}, we applied our own protocol excluding the ChEMBL subset. As shown in Appendix Figures \ref{fig: ood1} and \ref{fig: ood2}, MolReasoner outperforms Qwen-2.5-7b-Instruct on OOD tasks and significantly surpasses Mol-Instructions on the molecular OOD benchmark, demonstrating that our combination of SFT, RL, and expert-guided CoT data improves generalization, making MolReasoner more applicable in real-world, unseen scenarios.

\section{Conclusion}
% In this paper, we introduce \textbf{MolReasoner}, a two-stage training framework designed to advance large language models (LLMs) from memorization toward molecular reasoning. 
% By effectively combining supervised fine-tuning (Mol-SFT) and reinforcement learning (Mol-RL), MolReasoner equips LLMs with interpretable reasoning capabilities specifically tailored for molecular translation tasks. 
% Experimental evaluations demonstrate that MolReasoner significantly enhances the accuracy, interpretability, and structural comprehension of molecule-to-text and text-to-molecule translations. 
% Our findings highlight the importance of incorporating explicit reasoning into training paradigms, hopefully paving the way for advanced LLM applications in drug discovery, chemical synthesis, and materials science.
In this paper, we introduce MolReasoner, a two-stage framework that shifts large language models from memorization to effective reasoning in molecular tasks. 
Experimental evaluations demonstrate that MolReasoner significantly enhances the accuracy, interpretability, and structural comprehension of molecule-to-text and text-to-molecule translations.

\section{Limitations}
% Despite the strong empirical gains of MolReasoner, several limitations remain. First, our reliance on synthetic CoT rationales generated by GPT-4o may introduce biases and errors, and we lack a calibrated confidence estimate for its reasoning chains. Second, the reward functions, while effective at improving structural similarity and validity, do not account for properties such as synthetic accessibility or 3D conformational feasibility; more comprehensive and experimentally grounded evaluations are needed. Finally, the two-stage fine-tuning and on-policy reinforcement learning pipeline incurs significant computational cost, which may limit scalability to larger models and molecule libraries; optimizing for efficiency will be a key direction for our future work.

Despite MolReasoner's strong empirical results, some limitations remain. First, the synthetic CoT rationales from GPT-4o may introduce biases and errors, and we lack calibrated confidence estimates for these reasoning chains. Second, the reward functions focus on structural similarity and validity but do not address properties like synthetic accessibility or 3D conformational feasibility, necessitating further evaluations. Lastly, the two-stage fine-tuning and on-policy RL pipeline is computationally expensive, limiting scalability to larger models and molecule libraries. Future work will focus on improving efficiency.

\section{Ethics Statement}
In developing MolReasoner, we prioritized ethical considerations to ensure the responsible use of our models and methodologies. First, this research does not involve human subjects, and all datasets used (e.g., ChEBI-20) are publicly available and copyright-compliant. We applied strict data filtering to guarantee chemical validity and minimize the risk of introducing biased or misleading molecule–text pairs. Nevertheless, we acknowledge that biases inherent in benchmark datasets (e.g., underrepresentation of certain molecular families) may propagate into the model’s outputs. We adhere to all relevant legal and ethical research guidelines, including respecting open-source licenses during dataset construction and providing comprehensive model documentation. Our work is conducted with a strong commitment to research integrity, ensuring that our contributions remain beneficial to the scientific community and the AI for Science domain while addressing the ethical responsibilities associated with molecular AI technologies.

\bibliography{acl2026_conference}
\appendix
\clearpage
%\onecolumn
\newpage
\section{Reproducibility Statement}
We have made extensive efforts to ensure the reproducibility of our work. The proposed MolReasoner training pipeline is described in detail in the main text and appendix, including the Mol-SFT and Mol-RL stages. Details of dataset construction, reward design, and evaluation metrics are provided in the main text and appendix. We provide the code in the supplement material. Through these efforts, we aim to enable future research to reliably reproduce, validate, and extend our findings.

\section{The Use of Large Language Models}\label{appx: llm}

The authors declare that the human authors are the sole contributors to this work. This paper was written and edited exclusively by the authors. Large Language Models (LLMs) were used solely as a general-purpose tool to aid in the writing and editing process. Specifically, an LLM was utilized for:
\begin{itemize}
    \item \textbf{Text Polishing}: Improving the grammar, syntax, and flow of certain paragraphs to enhance overall readability.
    \item \textbf{Data Augmentation and Model Evaluation}: Assisting in generating synthetic data for supervised fine-tuning and served as a tool for evaluating model responses. Detailed prompts and methods for this data synthesis are provided in the main paper and in the 'Prompts' section of the appendix. All data generated or evaluated with the assistance of the LLM were thoroughly verified by the authors.
\end{itemize}
The authors have reviewed, edited, and verified all content generated or augmented by the LLM and take full responsibility for the entire contents of the paper. The use of LLMs does not constitute authorship.
\section{Related Work}
\label{appx: relatedwork}
In this section, we provide a review of literature related to molecular language models and large reasoning models.
\subsection{Molecular Language Models}
Early approaches to molecular understanding represent molecules as 1D sequences.
KV-PLM~\cite{kvplm} leverages SMILES~\cite{smiles} to represent molecules and employs a masked-language-modeling objective for pretraining on biomedical texts.
MolT5~\cite{molt5}, based on T5~\cite{T5}, is specifically designed for molecular translation tasks.
More recently, LlaSMol~\cite{llasmol} fine-tunes a suite of open-source LLMs on self-curated molecular instruction datasets. 
Mol-Instructions \cite{mol-instruction} adopts the SELFIES \cite{selfies} molecular descriptor and introduces a dedicated instruction dataset for biomolecular research.
The introduction of molecular graph encoders has led to the development of multimodal molecular language models.
MoMu~\cite{momu} and MoleculeSTM~\cite{stm} employ cross-modal contrastive learning to bridge the representation spaces of molecular graphs and text.
MolCA~\cite{molca} combines SMILES with 2D molecular representations for molecule-to-text generation.
More recent work, such as 3D-MoLM~\cite{3d-molm} and BioT5+~\cite{biot5+}, incorporates 3D molecular structures to enhance LLMs' ability to model molecular understanding.
Despite these advancements, these models remain limited in their reasoning capabilities due to the absence of Chain-of-Thought (CoT)~\cite{cot} fine-tuning. 
This gap restricts their performance in tasks requiring complex molecular understanding and reasoning, thereby limiting their practical utility in more demanding biomedical applications. While Mol-LLaMA~\cite{Mol-LLaMA} introduces a multi-modal alignment approach leveraging hierarchical reasoning across structural, chemical, and biological levels, its reliance on the SFT method makes it prone to overfitting, raising concerns about "reasoning fidelity" and "generation accuracy".

\subsection{Large Reasoning Models}
Recent advancements have led to the emergence of Large Reasoning Models (LRMs)~\cite{o1,gemini,deepseek}, which extend the capabilities of traditional LLMs by enabling deliberative, multi-step reasoning. 
These models distinguish themselves through the explicit representation of reasoning processes, which is crucial for tackling complex tasks requiring structured problem-solving.
The development of LRMs is closely tied to policy optimization techniques for model alignment, with the canonical approach being Reinforcement Learning from Human Feedback (RLHF)~\cite{rlhf} with Proximal Policy Optimization (PPO)~\cite{ppo}.
However, the computational demands and complexity of this approach, which involves managing multiple models (\ie policy, reference, reward, and critic), have driven the exploration of more efficient alternatives.
One such alternative is Group Relative Policy Optimization (GRPO), introduced in the training of DeepSeekMath~\cite{deepseekmath}.
GRPO is a variant of PPO that eliminates the need for a separate critic network, thereby reducing both memory and computational overhead. 
The efficiency and versatility of GRPO have enabled its application in a variety of high-stakes domains beyond its initial focus on mathematics, including puzzles~\cite{logic-r1}, medicine~\cite{med-r1}, and finance~\cite{fin-r1}.
Despite the progress in LRM development, their application to molecular tasks remains relatively underexplored.
The recently proposed ether0 ~\cite{ether0}, based on GRPO, has been trained on a large number of chemical problems and performs well on closed-set prediction tasks. However, this method relies on binary rewards, overlooking the importance of molecular fragment structure, and is prone to generating structural hallucinations in open-set generation tasks.
% presenting a promising avenue for this paper.

\section{Molecule-Text Translation}\label{appx: task_intro}
Advancing the use of LLMs in molecular science requires moving beyond structural representations toward explicit structure-level reasoning grounded in natural language.
Such reasoning abilities are essential for interpreting molecular semantics, inferring molecular functions, and generating chemically plausible structures from textual inputs.
To systematically evaluate these capabilities,  researchers~\cite{molt5} introduced the \textit{molecule-text translation} task, designed to assess a model’s proficiency in aligning and reasoning between molecular representations and natural language descriptions. 
This task comprises two complementary sub-tasks:
\begin{itemize}[leftmargin=*]
\item \textbf{Molecule Captioning}: Given a molecular descriptor (\eg SMILES~\cite{smiles}, SELFIES~\cite{selfies}, or IUPAC name~\cite{iupac}), the objective is to generate a coherent natural-language description capturing structural characteristics, functional roles, and potential applications of the molecule. 
This evaluates the model's ability to interpret molecular structures and abstract their semantics into text.
\item \textbf{Text-based de novo Molecule Generation}: Conversely, this task requires the model to generate valid molecular descriptors from natural language descriptions, testing its capacity to map textual semantics to relevant chemical motifs, and produce syntactically and chemically valid molecular structures.
\end{itemize}
Previous studies~\cite{galactica,molt5} uses SMILES for molecule representation, its grammar and token order sensitivity often result in invalid or chemically implausible outputs. 
primarily employed SMILES for molecular representation. However, the grammar and token-order sensitivity inherent in SMILES frequently lead to invalid or chemically implausible outputs. To overcome this limitation, we adopt SELFIES representations, following the Mol-Instructions framework~\cite{mol-instruction}. 
SELFIES ensures chemical validity by construction, eliminating common structural errors such as mismatched parentheses, invalid atomic symbols, and illogical branching patterns.

\subsection{Group Relative Policy Optimization}\label{appx: grpo}
The core innovation of Group Relative Policy Optimization (GRPO)~\cite{deepseekmath} lies in its group-based redefinition of the advantage function.
In contrast to PPO, GRPO removes the value function and estimates advantages relative to a sampled response group.
Specifically, given a question-answer pair $(q,a)$, the old policy $\pi_{\theta_\text{old}}$ samples a group of $G$ responses $\{ o_i\}_{i=1}^G$. 
The advantage for the $i$-th response is computed by normalizing the corresponding group-level rewards $\{ R_i \}_{i=1}^G$:
\begin{equation}
\hat{A}_{i,t} = \frac{r_i - \text{mean}(\{R_i\}_{i=1}^G)}{\text{std}(\{R_i\}_{i=1}^G)}.
\end{equation}

Additionally, GRPO employs a clipped objective combined with a KL-divergence penalty term, defined as:
\begin{equation}
\begin{aligned}
\mathcal{J}_\text{GRPO}(\theta) &= 
\mathbb{E}_{(q,a)\sim \mathcal{D}, \{o_i\}_{i=1}^G \sim \pi_{\theta_\text{old}}(\cdot\mid q)} \Bigg[
\frac{1}{G}\sum_{i=1}^{G} \frac{1}{|o_i|} \sum_{t=1}^{|o_i|} \Bigg( \\
&\quad \min \Big( r_{i,t}(\theta) \hat{A}_{i,t},\ 
\text{clip} \left( r_{i,t}(\theta), 1 - \varepsilon, 1 + \varepsilon \right) \hat{A}_{i,t} \Big) \\
&\quad - \beta D_{\text{KL}}(\pi_{\theta} \parallel \pi_{\text{ref}}) 
\Bigg) \Bigg]
\end{aligned}
\label{eq:grpoloss}
\end{equation}
where the importance ratio $r_{i,t}(\theta)$ is given by:
\begin{equation}
    r_{i,t}(\theta)=\frac{\pi_{\theta}(o_{i,t} \mid q, o_{i,<t})}{\pi_{\theta_{\text{old}}}(o_{i,t} \mid q,o_{i,<t})}.
\end{equation}

\section{Metrics.} \label{appx: metrics}

(1) \textbf{Frag-J:} Measures the Jaccard similarity between the fragment sets of the predicted and reference molecules. Higher values indicate larger structural overlap.
(2) \textbf{Frag-R:} Assesses whether key structural fragments in the reference molecule are recalled in the prediction.
(3) \textbf{FG-Match (Functional Group Matching):} Computes the difference in the number of functional groups between the prediction and the reference using an exponential decay formulation; a higher score indicates greater similarity.

\section{Baseline and Training Setup.} \label{appx: trainingsetup}
For the baseline models,
we deliberately select baselines that are LLM-based models.
For prompt-based methods, we compare our model against leading general-purpose LLMs, including GPT-4o, GPT-4o-mini, Deepseek-r1-0528, Qwen2.5-7B-Instruct, DeepSeek-R1-Distill-Qwen-7B, LLaMA3.1-8B-Instruct, Qwen3-8B, Qwen2.5-32B-Instruct, LLaMA3.1-70B-Instruct, Qwen2.5-72B-Instruct, and LLaMA3 with MSR(10-shot). 
For fine-tuning approaches without explicit reasoning, we evaluate Mol-Instructions + LLaMA2/3 (without fine-tuning), Mol-LLaMA, and their fine-tuned versions: Mol-Instructions + LLaMA3 (ft) and LLaMA3 + MSR (ft). Also, we compare  chemical reasoning model ether0. 

We use Qwen2.5-7B-Instruct as the base model. During the warm-up stage, we fine-tune the model separately on CoT data for both tasks. The learning rate is set to $1\times10^{-5}$, with a total of 4 epochs, and a cosine learning rate scheduler is applied to stabilize training.
Following warm-up, we perform GRPO reinforcement learning. In this stage, the batch size is 256, the learning rate is $1\times10^{-7}$, and each sample generates 8 candidate outputs (rollouts). A temperature of 0.7 is applied to encourage output diversity. We train the models for 15 epochs to fully optimize generation performance across both tasks. All experiments are run on 8 Tesla-A100-80G GPUs. For molecule captioning and text-based de novo molecule generation, Mol-SFT requires approximately 1 GPU hour. Mol-RL requires approximately 1200 GPU hours.

\section{Prompts}\label{appx: gpt_cot}
In this section, we present the input prompts used for knowledge‑guided chain‑of‑thought data construction with GPT‑4o, and also the prompts used to evaluate the quality of the model responses. These prompts cover both the text-based de novo molecule generation and the molecule captioning tasks.
\onecolumn
\begin{myexample}{Molecule captioning task}{}
    You are a professional chemist. Given a molecule's \textbf{SELFIES} and \textbf{structural features}, and identified \textbf{functional fragments}, your task is to \textcolor{blue}{generate a natural, concise, and chemically accurate description of the molecule}. Do not reproduce or reference the original SELFIES string. Instead, decode it internally and abstractly summarize the key structural features it represents.\\
        \textbf{SELFIES}: \\
       {%
  \ttfamily
  [O]\allowbreak
  [=C]\allowbreak
  [Branch1]\allowbreak
  [C]\allowbreak
  [O-1]\allowbreak
  [C]\allowbreak
  [C]\allowbreak
  [S]\allowbreak
  [S]\allowbreak
  [C]\allowbreak
  [Ring1]\allowbreak
  [Branch1]
}\\
        \textbf{Structural Info}: \\
        1. The molecule has 1 ring(s), including 0 aromatic ring(s).\\
        2. The molecular weight is approximately 149.22 g/mol.\\
        \textbf{Fragments}: \\
        {\ttfamily
    {<|O=C[O-]|>}\allowbreak
    {<|SS|>}\allowbreak
    {<|C030|>}\allowbreak
    {<|C020|>}\allowbreak
    {<|C020|>}
  }\\
    Please provide a step-by-step analysis that explains how you would interpret this molecular structure and summarize it into a concise and chemically accurate natural language description. Let's think step by step and return the final answer in \textcolor{pink}{\textless{}answer\textgreater{} The molecule is a monocarboxylic acid anion and a member of dithiolanes. It is a conjugate base of an asparagusic acid. It derives from a hydride of a 1,2-dithiolane. \textless{}/answer\textgreater{}} tags.
\end{myexample}

\begin{myexample}{Text-based molecule generation task}{}
 You are a professional biochemist designing molecular structures. Given the \textbf{molecular description}, basic \textbf{structural information}, and identified \textbf{fragments}. Your goal is \textcolor{blue}{not to re-predict} the molecule's SELFIES, but to \textcolor{blue}{generate a logical, chemically sound reasoning chain} that explains how one could deduce or construct this structure based on the given information.\\
\textbf{Description}: \\
1. The molecule is an aldehyde that is thiphene substituted by a formyl group at position.\\
2. It has a role as a metabolite. It is a member of thiophenes and an aldehyde\\
\textbf{Structural Info}:\\1. The molecule has 1 ring(s), including 1 aromatic ring(s).\\2. The molecular weight is approximately 112.15 g/mol.\\
\textbf{Fragments}:\\
{\ttfamily
    {<|C=O|>}\allowbreak
    {<|c1ccsc1|>}
  }\\
Please provide a step-by-step \textcolor{blue}{molecular reasoning chain} that explains how you would reconstruct or deduce the molecular structure. Let's think step by step and return the final answer in \textcolor{pink}{\textless{}answer\textgreater{} {%
  \ttfamily
  [O]\allowbreak
  [=C]\allowbreak
  [C]\allowbreak
  [=C]\allowbreak
  [C]\allowbreak
  [=C]\allowbreak
  [S]\allowbreak
  [Ring1]\allowbreak
  [Branch1]
}\textless{}/answer\textgreater{}} tags.
\end{myexample}
\begin{myexample}{Example for quality evaluation of the model responses}{}
You are a professional chemist and an expert evaluator. Your task is to assess and score a molecular description (Chain of Thought, or CoT) generated by an AI model, based on specific criteria. Your evaluation must be strict, objective, and consistent with the provided examples.
Please score the CoT on a scale of 1-5 across three key dimensions:
\section*{1. Clarity of Logic \& Insightfulness}
This criterion evaluates whether the CoT's reasoning process is coherent and if it provides valuable chemical insights.
\begin{itemize}
\item \textbf{5 (Outstanding)}: The reasoning is exceptionally clear, with a flawless logical progression. The CoT provides profound chemical insights beyond simple facts, demonstrating expert-level thought.
\item \textbf{4 (Excellent)}: The logical flow is very clear and the analysis is systematic. It provides accurate insights but may lack particularly novel or deep observations.
\item \textbf{3 (Acceptable)}: The reasoning is mostly clear, but there may be minor logical jumps. The insights provided are correct but basic.
\item \textbf{2 (Lacking)}: The reasoning is disorganized or difficult to follow. The connection between analysis and conclusions is weak.
\item \textbf{1 (Poor)}: There is no recognizable logic or insight. The CoT is a disorganized list of facts with no meaningful analysis.
\end{itemize}

\section*{2. Factual Correctness}
This criterion evaluates the accuracy of all factual statements within the CoT.
\begin{itemize}
\item \textbf{5 (Completely Accurate)}: All chemical statements, nomenclature, and factual details are entirely correct with no inaccuracies.
\item \textbf{4 (Minor Errors)}: Contains one or two subtle, inconsequential errors that do not affect the overall conclusion.
\item \textbf{3 (Partially Accurate)}: Contains a few identifiable factual errors or inaccuracies that do not fundamentally break the reasoning.
\item \textbf{2 (Multiple Errors)}: Contains several clear and misleading factual errors.
\item \textbf{1 (Severely Flawed)}: Riddled with serious factual errors, making the entire analysis untrustworthy.
\end{itemize}

\section*{3. Redundancy \& Conciseness}
This criterion measures the efficiency of the CoT. A high-quality CoT should contain only necessary steps, avoiding repetition.
\begin{itemize}
\item \textbf{5 (Extremely Concise)}: Every step in the CoT is essential. There are no redundant sentences or repeated analyses; the text is efficient and to the point.
\item \textbf{4 (Concise)}: Most steps are necessary. There may be one or two sentences that could be trimmed, but the overall text is not redundant.
\item \textbf{3 (Acceptable)}: Contains some redundant information that could be merged or removed, but the overall structure remains clear.
\item \textbf{2 (Verbose)}: Contains repetitive analysis or unnecessary information that makes the text feel bloated.
\item \textbf{1 (Extremely Verbose)}: The text is filled with a large amount of repetitive or irrelevant content, making it difficult to read.
\end{itemize}

\hrule
\vspace{12pt}

\textbf{Important Note: Adjusting for Scoring Bias}

Please be aware that human experts tend to be more lenient and generous in their scoring than you. When a CoT is of high quality, experts often assign scores of \textbf{4 or 5} even if there are minor imperfections. In contrast, your current scoring may be too conservative.

When performing your evaluation, please adjust your internal scoring scale to align with this more generous, expert-like style. For high-quality CoTs, \textbf{do not hesitate to assign 4s and 5s}.

\hrule
\vspace{12pt}

Your final evaluation must be returned as a \textbf{JSON object}, and it must contain only this JSON. The JSON must include the following keys:
\begin{itemize}
\item \texttt{clarity\_score}: The score for Clarity of Logic \& Insightfulness (1-5)
\item \texttt{factual\_score}: The score for Factual Correctness (1-5)
\item \texttt{conciseness\_score}: The score for Redundancy \& Conciseness (1-5)
\end{itemize}
\section*{Few-Shot Examples}

\textbf{Follow these examples strictly, mimicking the format and scoring style:}\\
\textbf{Example 1:}\\
...\\
\textbf{Human expert's score:}
\{
  "clarity score": 5,
  "factual score": 4,
  "conciseness score": 4,
\}\\
\textbf{Example 2:}\\
...\\
\textbf{Human expert's score:}
\{
  "clarity score": 3,
  "factual score": 5,
  "conciseness score": 4,
\}\\
\textbf{Example 3:}\\
...\\
\textbf{Human expert's score:}
\{
  "clarity score": 4,
  "factual score": 3,
  "conciseness score": 5,
\}\\
Please rate the following model response:

\end{myexample}
\twocolumn
% \section{Ablation Study on Reward Design}

% \section{Ablation Study on CoT}

% \section{Ablation Study on SMILES and SELFIES}

\section{Success and Failure Cases}
\label{appx: cases}
In this section, we present representative success and failure cases for both molecule generation and molecule captioning tasks, with some basic analysis of their respective error modes. Table \ref{tab:molecule generation case 1} and \ref{tab:molecule generation case 2} show the model responses for molecule generation,  
while Table \ref{tab:molecule caption case 1} and \ref{tab:molecule caption case 2} show those for molecule captioning.
\onecolumn
\begin{table*}[t]
\caption{\textbf{Case \uppercase\expandafter{\romannumeral 1} used to illustrate the model responses of the molecule generation task.} The top block shows the prompt used to test the models, and the following blocks show the response of prompt-based method(Qwen2.5-72B-Instruct), the model after fine-tuning without explicit reasoning and our method.}
\label{tab:molecule generation case 1}
\centering
\begin{minipage}{\columnwidth}\vspace{0mm}    \centering
\begin{tcolorbox} 
    \centering
      \small
    \begin{tabular}{p{\columnwidth} c}
    {\bf Prompt}  & \\
You are a professional biochemist designing molecular structures. Please generate the molecular structure (SELFIES) based on the following description. Please \textbf{think step by step} and return the final answer in \textcolor{purple}{\textless{}answer\textgreater{} ... \textless{}/answer\textgreater{}} tags. & \\
The molecule is a phosphatidylcholine 32:0 in which the 1- and 2-acyl groups are specified as hexadecanoyl (palmitoyl). A synthetic phospholipid used in liposomes and lipid bilayers to study biological membranes. It is also a major constituent of pulmonary surfactants. It has a role as a surfactant and a mouse metabolite. It is a phosphatidylcholine 32:0 and a 1-acyl-2-hexadecanoyl-sn-glycero-3-phosphocholine. It derives from a hexadecanoic acid. & \\

\textcolor{blue}{\textbf{Ground truth}: \textnormal{[C][C][C][C][C][C][C][C][C][C][C][C][C][C][C][C][=Branch1][C][=O][O][C]
[C@H1][Branch2][Ring1][Branch1][C][O][P][=Branch1][C][=O][Branch1][C][O-1][O][C][C][N+1][Branch1]
[C][C][Branch1][C][C][C][O][C][=Branch1][C][=O][C][C]
[C][C][C][C][C][C][C][C][C][C][C][C][C]}}

    \hrulefill & \\
    {\bf Prompt-based method: Qwen2.5-72B-Instruct}  & \\
To generate the SELFIES (SELF-referencING mOLECULAR stringS) for the phosphatidylcholine 32:0 with 1- and 2-acyl groups specified as hexadecanoyl (palmitoyl), we need to break down the structure step by step. & \\

Step-by-Step Breakdown & \\

1. Hexadecanoyl (Palmitoyl) Chain: Hexadecanoyl is a 16-carbon saturated fatty acid. SELFIES for hexadecanoyl: [C][C][C][C][C][C][C][C][C][C][C][C][C][C][C][C][O][=O] & \\

2. Glycerol Backbone: Glycerol is a three-carbon alcohol. SELFIES for glycerol: [C][O][C][O][C][O] & \\

3. Phosphate Group: The phosphate group is attached to the third carbon of glycerol. SELFIES for phosphate: [P][O][O][O][O] & \\

4. Choline Head Group: Choline is a quaternary ammonium compound.  SELFIES for choline: [N][C][C][C][O][C][C][C][C][C][C][C]...(approx. 100 [C])[C][C]& \\
\textcolor{red}{\textbf{(Hallucinated structure. Repetitive fragments. Format not followed.)}} & \\
    \hrulefill & \\

    {\bf Model after fine-tuning Without Explicit Reasoning}  & \\
   \textnormal{[C][C][C][C][C][C][C][C][C][C][C][C][C][C][C][C][=Branch1][C][=O][O][C][C@H1[Branch2]
   [Ring1][Branch1][C][O][P][=Branch1][C][=O][Branch1][C][O][O][C][C][N+1][Branch1][C][C]
   [Branch1][C][C][C][O][C][=Branch1][C]} & \\
\textcolor{red}{\textbf{(Valid but incorrect molecule. No reasoning. Format not followed.)}} & \\

    \hrulefill & \\
    {\bf Our method}  & \\
    \textless think\textgreater{} & \\
To deduce the molecular structure of the phosphatidylcholine 32:0 with specified acyl groups, we can follow a logical reasoning chain based on the provided description, structural information, and fragments. Here’s a step-by-step breakdown: & \\
\textbf{1. Identify the Core Structure}: Phosphatidylcholine is a type of ... & \\
\textbf{2. Determine the Glycerol Backbone}: The glycerol backbone is a three-carbon molecule with ... & \\
\textbf{3. Acyl Groups}: The description specifies that the 1- and 2-acyl groups are hexadecanoyl (palmitoyl). Hexadecanoyl refers to a fatty acid with 16 carbon atoms (C16). Therefore, ... & \\
\textbf{4. Phosphocholine Head Group}: The third carbon of the glycerol is attached to a phosphocholine group. This group ... & \\
… (omitted structural construction and molecular weight calculation steps) & \\
\textbf{8. Use the Provided Fragments}: The fragments given in the description can be used to construct the molecular structure. For example, the fragment ... & \\
… (omitted rest of structural deducing reasoning, less central to SELFIES construction) & \\
The final answer, representing the constructed molecular structure, is: & \\
\textless answer\textgreater \textnormal{\textbf{[C][C][C][C][C][C][C][C][C][C][C][C][C][C][C][C][=Branch1][C][=O][O][C]
[C@H1][Branch2][Ring1][Branch1][C][O][P][=Branch1][C][=O][Branch1][C][O-1][O][C][C][N+1][Branch1][C][C][Branch1][C][C][C][O][C][=Branch1][C][=O][C][C]
[C][C][C][C][C][C][C][C][C][C][C][C][C]}}\textless /answer\textgreater & \\
\textcolor{red}{\textbf{(Clear reasoning. Valid and correct molecule output. Desired format.)}} & \\

    \end{tabular}
\end{tcolorbox}

\end{minipage}
% \vspace{-12pt}
\end{table*}
\twocolumn

\onecolumn
\begin{table*}[t]
\caption{\textbf{Case \uppercase\expandafter{\romannumeral 2} used to illustrate the model responses of the molecule generation task.} The top block shows the prompt used to test the models, and the following blocks show the response of prompt-based method(DeepSeek-R1-Distill-Qwen-7B), the model after fine-tuning without explicit reasoning and our method.}
\label{tab:molecule generation case 2}
\centering
\begin{minipage}{\columnwidth}\vspace{0mm}    \centering
\begin{tcolorbox} 
    \centering
      \small
    \begin{tabular}{p{\columnwidth} c}
    {\bf Prompt}  & \\
You are a professional biochemist designing molecular structures. Please generate the molecular structure (SELFIES) based on the following description. Please \textbf{think step by step} and return the final answer in \textcolor{purple}{\textless{}answer\textgreater{} ... \textless{}/answer\textgreater{}} tags. & \\
The molecule is a 1-O-acyl-sn-glycero-3-phosphocholine in which the acyl group is specified as caproyl (hexanoyl). It is a 1-O-acyl-sn-glycero-3-phosphocholine, a hexanoate ester and a lysophosphatidylcholine 6:0. & \\

\textcolor{blue}{\textbf{Ground truth}: \textnormal[C][C][C][C][C][C][=Branch1][C][=O][O][C][C@@H1][Branch1][C][O][C][O][P]
[=Branch1][C][=O][Branch1][C][O-1][O][C][C][N+1][Branch1][C][C][Branch1][C][C][C]}

    \hrulefill & \\
    {\bf Prompt-based method: DeepSeek-R1-Distill-Qwen-7B}  & \\
\textless think\textgreater{}& \\
Okay, so I need to figure out how to construct the SELFIES notation for this molecule. Let me start by understanding the structure. The molecule is ... But in SELFIES, it's represented as \textnormal{[C][C@H1][C@@H1][Branch2]}. \textcolor{pink}{Hmm}, maybe I should break it down....(another round of reasoning)The hexanoyl would be a chain of six carbons with an ester group, so maybe \textnormal{[C][C][C](O)=O} connected to the hexanoyl.& \\
\textcolor{pink}{Wait}, perhaps it's better to structure it step by step. The choline part is ...(reasoning to construct the SELFIES)I think the correct SELFIES would be: \textnormal{[C][C@H1][C@@H1][Branch2]O[C][C][C](O)=O[C][C][C]}. \textcolor{pink}{Wait}, that might not be accurate. Let me make sure.Alternatively, maybe it's \textnormal{[C][C@H1][C@@H1][Branch2]O[C][C][C](O)=O[C][C][C]}. No, that doesn't seem right because ... Putting it all together, the SELFIES should be: [C][C@H1][C@@H1][Branch2]O[C][C][C](O)=O[C][C][C]. \textcolor{pink}{Wait}, no, the hexanoyl is ... \textcolor{pink}{Hmm, I'm getting confused}. Let me try to write it step by step.1. Choline: [C][C@H1][C@@H1][Branch2]... So, perhaps: \textnormal{[C][C@H1][C@@H1][Branch2]O[C][C][C](O)=O[C][C][C][C][C][C]} &\\
\textcolor{pink}{Wait}, that still doesn't look right. Maybe I should ... Putting it all together, the SELFIES would be:[C][C@H1][C@@H1][Branch2]O[C][C][C](O)=O[C][C][C][C][C][C]& \\ \textcolor{pink}{Wait, no,} the hexanoyl should be... The glycerol's C2 has the hexanoyl: [C][C][C](O)=O.& \\
\textcolor{pink}{Wait}, perhaps SELFIES is:\textnormal{[C][C@H1][C@@H1][Branch2]O[C][C][C](O)=O[C][C][C][C][C][C]}.&\\
\textcolor{pink}{But I'm not confident. Maybe} I should represent the glycerol as  ... Putting it all together, the SELFIES would be:\textnormal{[C][C@H1][C@@H1][Branch2]O[C][C][C](O)=O[C][C][C][C][C][C]}. \textcolor{pink}{Wait}, that seems too long. & \\
\textcolor{red}{\textbf{(Frequent self-doubt. Format not followed.)}} & \\

    \hrulefill & \\

    {\bf Model after fine-tuning Without Explicit Reasoning}  & \\
   \seqsplit{[C][C][C][C][C][C][=Branch1][C][=O][O][C][C@H1][Branch2][Ring1][Branch1][C][O][P][=Branch1][C][=O][Branch1][C][O][O][C][C][N+1][Branch1][C][C][Branch1][C][C][C][O]} & \\
\textcolor{red}{\textbf{(Lack of Chain-of-Thought. Format not followed. Generally accurate yet deviates subtly from the ground truth.)}} & \\

    \hrulefill & \\

    {\bf Our method}  & \\
    \textless think\textgreater{} & \\
To deduce the molecular structure of 1-O-acyl-sn-glycero-3-phosphocholine with a caproyl (hexanoyl) acyl group, we can follow a logical reasoning chain based on the provided description, structural information, and fragments. Here’s a step-by-step breakdown:&\\
1. \textbf{Identify the Core Structure}: The molecule is a type of phospholipid, specifically ...&\\
2. \textbf{Acyl Group}: The acyl group is specified as caproyl, which is a straight-chain fatty acid with six carbon atoms (hexanoate). This means the acyl group is \textnormal{-C(=O)CH2CH2CH2CH2CH3}.&\\
3. \textbf{Glycerol Backbone}: The glycerol backbone consists of three carbon atoms, where ...&\\
4. \textbf{Phosphocholine Group}: The phosphocholine group consists of a phosphorus atom bonded to ...&\\
5. \textbf{Construct the Molecular Structure}: Start with the glycerol backbone: C1 is attached to ...&\\
6. \textbf{Verify the Molecular Weight}: The molecular weight of the constructed molecule should be ...&\\
7. \textbf{Check the Chemical Connectivity}: Ensure that all atoms are connected correctly and ...&\\
8. \textbf{Finalize the Structure}: Once the structure is constructed and verified, it can be represented in a standardized format, such as SELFIES or a similar notation.&\\

The final answer, representing the constructed molecular structure, is: &\\\textless answer\textgreater{}\seqsplit{[C][C][C][C][C][C][=Branch1][C][=O][O][C][C@@H1][Branch1][C][O][C][O][P][=Branch1][C][=O][Branch1][C][O-1][O][C][C][N+1][Branch1][C][C][Branch1][C][C][C]} \textless /answer\textgreater{}&\\
\textcolor{red}{\textbf{(Clear and transparent reasoning. Valid and correct molecule output. Desired format.)}} & \\
    \end{tabular}
\end{tcolorbox}

\end{minipage}
% \vspace{-12pt}
\end{table*}

\begin{table*}[t]
\caption{\textbf{Case \uppercase\expandafter{\romannumeral 1} used to illustrate the model responses of the molecule captioning task.} The top block shows the prompt used to test the models, and the following blocks show the response of prompt-based method(GPT-4O), the model after fine-tuning without explicit reasoning and our method.}
\label{tab:molecule caption case 1}
\centering
\begin{minipage}{\columnwidth}\vspace{0mm}    \centering
\begin{tcolorbox} 
    \centering
      \small
    \begin{tabular}{p{\columnwidth} c}
    {\bf Prompt}  & \\

You are a professional chemist. Your task is to generate a natural, concise, and chemically accurate description of a given molecule.  Please provide a \textbf{step-by-step} analysis explaining how you interpret the molecular structure, identify key features and functional groups, and summarize it into a clear and informative description. Think step by step, and your final answer \textbf{must} be returned in the format: \textcolor{red}{\textless{}answer\textgreater{}} ... \textless{}/answer\textgreater{}. For example:&\\
\textless{}answer\textgreater{}The molecule is an epoxy(hydroxy)icosatrienoate that is the conjugate base of 11-hydroxy-(14R,15S)-epoxy-(5Z,8Z,12E)-icosatrienoic acid, obtained by deprotonation of the carboxy group; major species at pH 7.3. It is a conjugate base of an 11-hydroxy-(14R,15S)-epoxy-(5Z,8Z,12E)-icosatrienoic acid.\textless{}/answer\textgreater{} &\\

\textnormal{[C][S][C][C][C@H1][Branch1][C][NH3+1][C][=Branch1][C][=O][N][C@@H1][Branch1]
[C][C][C][=Branch1][C][=O][O-1]} &\\

\textcolor{blue}{\textbf{Label}: The molecule is a peptide zwitterion obtained by transfer of a proton from the carboxy to the amino terminus of Met-Ala. It is a tautomer of a Met-Ala.}

    \hrulefill & \\

    {\bf Prompt-based method: GPT-4o}  & \\
To interpret and generate a precise description of the given molecule from its SELFIES notation, follow these analytical steps: &\\
Step 1: Understand SELFIES Notation SELFIES (Self-referencing Embedded Strings) is a high-level representation of molecular structures. Each character or symbol corresponds to ... &\\
Step 2: Parse and Identify Atoms \textnormal{[C]}: Represents carbon atoms. \textnormal{[S]}: Represents a sulfur atom. \textnormal{[C@H1]}: A chiral carbon ... &\\ 
Step 3: Build the Backbone: By interpreting the sequence, the molecule exhibits a linear carbon chain with branching points, functional groups (amino and carboxylate), and stereochemical centers.&\\ 
Step 4: Identify Functional Groups and Stereochemistry: An amino group \textnormal{[NH3+1]} exists as a protonated NH3+. Two carbonyl groups (=O) are present one possibly as part of an amide and the other as a carboxylate. The sulfur atom \textnormal{[S]} suggests a thioether or thioamide moiety.&\\
Step 5: Summarize the Molecule: Integrating stereochemistry, charges, and functional groups results in the clear identification of the compound.&\\
\textless{}answer\textgreater{} The molecule is a sulfinyl-dipeptide derivative containing a protonated amino group at one terminus and a negatively charged carboxylate group at the other. It includes a sulfur atom within the backbone structure, flanked by two stereogenic centers marked with opposing stereochemistry (R and S configurations). The carbon skeleton features both an amide linkage and carboxylate functionality, consistent with its zwitterionic form at physiological pH (7.3).\textless{}/answer\textgreater{} & \\
\textcolor{red}{\textbf{(Inaccurate SELFIES parsing. Overgeneralized functional group assignments. Unsupported structural assumptions.)}} & \\

    \hrulefill & \\

    {\bf Model after fine-tuning Without Explicit Reasoning}  & \\
   The molecule is zwitterionic form of N(6),N(6),N(6)-trimethyl-L-lysine arising from transfer of a proton from the carboxy to the amino group; major species at pH 7.3 It is a tautomer of a N(6),N(6),N(6)-trimethyl-L-lysine. & \\
\textcolor{red}{\textbf{(Lack of Chain-of-Thought. Format not followed.)}} & \\

    \hrulefill & \\

    {\bf Our method}  & \\
    \textless think\textgreater{} & \\
To interpret the provided molecular structure based on the SELFIES, structural information, and functional fragments, we can follow these steps:&\\
\textbf{1. Identify the Core Structure}: The molecule appears to be a peptide, specifically a dipeptide, consisting of ...&\\
\textbf{2. Analyze the Functional Groups}: The key functional groups include:  An amino group (NH3+), which is ...&\\
\textbf{3. Examine the Amino Acid Residues}: The structure includes L-methionine and L-alanine: L-methionine contributes ...&\\
\textbf{4. Consider the Zwitterionic Nature}: At a pH of 7.3, the amino group is likely ...&\\
\textbf{5. Summarize the Findings:} Based on the analysis, we can conclude that the molecule is a ...&\\
Now, we can formulate a concise and chemically accurate description of the molecule:&\\
\textless answer\textgreater The molecule is a peptide zwitterion obtained by transfer of a proton from the carboxy to the amino group of Met-Ala; major species at pH 7.3. It is a tautomer of a Met-Ala. \textless /answer\textgreater & \\
\textcolor{red}{\textbf{(Clear reasoning. Valid and correct caption output. Desired format.)}} & \\
    \end{tabular}
\end{tcolorbox}

\end{minipage}
% \vspace{-12pt}
\end{table*}

\begin{table*}[t]
\caption{\textbf{Case \uppercase\expandafter{\romannumeral 2} used to illustrate the model responses of the molecule captioning task.} The top block shows the prompt used to test the models, and the following blocks show the response of prompt-based method(Qwen2.5-72B-Instruct), the model after fine-tuning without explicit reasoning and our method.}
\label{tab:molecule caption case 2}
\centering
\begin{minipage}{\columnwidth}\vspace{0mm}    \centering
\begin{tcolorbox} 
    \centering
      \small
    \begin{tabular}{p{\columnwidth} c}
    {\bf Prompt}  & \\
You are a professional chemist. Your task is to generate a natural, concise, and chemically accurate description of a given molecule. Please provide a step-by-step analysis explaining how you interpret the molecular structure, identify key features and functional groups, and summarize it into a clear and informative description. Think step by step, and your final answer must be returned in the format: \textless{}answer\textgreater{} ... \textless{}/answer\textgreater{}. For example: &\\

\textless{}answer\textgreater{}The molecule is an epoxy(hydroxy)icosatrienoate that is the conjugate base of 11-hydroxy-(14R,15S)-epoxy-(5Z,8Z,12E)-icosatrienoic acid, obtained by deprotonation of the carboxy group; major species at pH 7.3. It is a conjugate base of an 11-hydroxy-(14R,15S)-epoxy-(5Z,8Z,12E)-icosatrienoic acid.\textless{}/answer\textgreater{}&\\

\seqsplit{[C][S][C][C][C@H1][Branch1][S][N][C][=Branch1][C][=O][C@@H1][Branch1][C][N][C][C][Branch1][C][C][C][C][=Branch1][C][=O][O]}

\textcolor{blue}{\textbf{Label}: The molecule is a dipeptide formed from L-leucine and L-methionine residues. It has a role as a metabolite. It derives from a L-leucine and a L-methionine.}

    \hrulefill & \\

{\bf Prompt-based method: Qwen2.5-72B-Instruct}  & \\
\textbf{Step-by-Step Analysis}: &\\
Step 1: Identify the Backbone Structure. The given structure is a linear chain of carbon atoms with several branches and functional groups. The main chain can be identified as a series of carbon atoms connected by single and double bonds. &\\
Step 2: Identify Functional Groups. Carbonyl Groups: There are two carbonyl groups (C=O) present in the structure. Amine Groups: There are two amine groups (N) ...&\\
Step 3: Determine Stereochemistry C@H1: Indicates a chiral center with the R configuration...&\\
Step 4: Summarize the Molecule. Combining the information from the steps above, we can describe the molecule as follows:&\\
\textless{}answer\textgreater{}The molecule is a branched, chiral compound with a linear carbon backbone. It contains two carbonyl groups, two amine groups, and sulfur atoms. The chiral centers are configured as (R) and (S) at the specified positions. The structure suggests a complex organic molecule with potential biological activity due to its functional groups and stereochemistry.\textless{}/answer\textgreater{}&\\
\textcolor{red}{\textbf{Misidentification of the molecule. Lack of treatment of charge states or protonation. Overly vague final description.}}&\\
    \hrulefill & \\

    {\bf Model after fine-tuning Without Explicit Reasoning}  & \\
   The molecule is a L-histidine derivative that is N(alpha)-acetyl-L-histidine substituted by a methyl group at position 5 on the imidazole ring. It has a role as a human urinary metabolite. It is a L-histidine derivative and a N(alpha)-acetyl-L-histidine. It is a conjugate acid of a N(alpha)-acetyl-N(epsilon)-methyl-L-histidine(1-). & \\
\textcolor{red}{\textbf{(Lack of Chain-of-Thought. Format not followed. Misidentification of molecule)}} & \\

    \hrulefill & \\

    {\bf Our method}  & \\
    \textless think\textgreater{} & \\
To generate a concise and chemically accurate description of the molecule based on the provided SELFIES, structural features, and functional fragments, we can follow these steps:&\\
1. \textbf{Identify the Core Structure}: The molecule appears to be a peptide, specifically a dipeptide, given the presence of amino acid fragments and the presence of an amine (N) and a carboxylic acid (C=O and O).&\\
2. \textbf{Analyze the Functional Groups}: The presence of sulfur (S) atoms suggests the inclusion of methionine, an amino acid containing a sulfur atom. The presence of amine (N) and carboxylic acid (C=O and O) groups confirms the peptide nature of the molecule.&\\
3. \textbf{Determine the Amino Acid Composition}: The fragments indicate the presence of L-leucine and L-methionine. L-leucine is recognized by the branched carbon structure, while L-methionine is identified by the sulfur atom.&\\
4. \textbf{Consider the Stereochemistry}: The notation (C@H1) and (C@@H1) indicates that there are chiral centers in the molecule, which is typical for amino acids.&\\
5. \textbf{Summarize the Findings}: Based on the analysis, we can summarize the key features of the molecule, including its composition, structure, and biological relevance.&\\
Now, we can formulate the final answer:&\\
\textless answer\textgreater The molecule is a dipeptide composed of L-leucine and L-methionine joined by a peptide linkage. It has a role as a metabolite. It derives from a L-leucine and a L-methionine. \textless /answer\textgreater & \\
\textcolor{red}{\textbf{(Clear reasoning. Valid and correct caption output. Desired format.)}} & \\
    \end{tabular}
\end{tcolorbox}

\end{minipage}
% \vspace{-12pt}
\end{table*}
\twocolumn
\section{Reasoning--Answer Consistency and Error Typology}
\label{app:reasoning_error_analysis}

This section provides an additional analysis of how MolReasoner and generic LLM baselines fail on the two benchmarks (\textsc{Molecule Captioning} and \textsc{Text-based de novo Molecule Generation}), with a particular focus on the relationship between chain-of-thought (CoT) reasoning quality and final answer correctness.

\textbf{Sampling and annotation protocol.}
For each task and each model, we randomly sampled 150 examples from the evaluation set. We first filtered out cases where both the CoT reasoning and the final answer are clearly correct according to task-specific criteria (defined below). From the remaining pool of error cases, we randomly selected 100 instances per model and task for detailed annotation. The annotations were produced with the help of GPT-5 under a model-specific rubric, followed by manual spot checks to ensure consistency.

Each error case is assigned two binary labels:
(i) answer correctness (\textsc{Answer Correct} vs.\ \textsc{Answer Wrong});
and (ii) reasoning quality (\textsc{Reasoning Correct} vs.\ \textsc{Reasoning Flawed}).
This yields three main categories among the error cases:
(1) correct reasoning but wrong answer;
(2) flawed reasoning but correct answer;
(3) both reasoning and answer wrong.
We further attach a fine-grained reasoning error type to each \textsc{Reasoning Flawed} case.

\textbf{Task-specific criteria for answer correctness.}
For \textsc{Molecule Captioning} (captioning), a prediction is labelled \textsc{Answer Correct} if it matches the gold description at three levels:
(1) the molecular class / scaffold (e.g.\ both describe a thiazolium salt instead of one being a steroid and the other a carbohydrate);
(2) the core functional groups and substitution pattern (e.g.\ N-methylation, para-hydroxy substitution, phosphate diester vs.\ carboxylate) without essential omissions or fabrications;
and (3) the acid--base form when the gold description explicitly focuses on conjugate acid/base.
Any serious mismatch on these dimensions (e.g.\ sulfate vs.\ carboxylic acid, conjugate base described as neutral acid) is counted as \textsc{Answer Wrong}.
For \textsc{Text-based de novo Molecule Generation} (generation), we decode the predicted SELFIES into a molecule and compare it with the reference structure. If the two molecules are topologically identical (canonical SMILES match), we label the answer as \textsc{Answer Correct}. In rare cases where stereochemical annotations differ but the underlying graph is identical, we still treat the answer as correct.

\textbf{Criteria for reasoning quality.}
We evaluate the CoT reasoning along four dimensions. Reasoning is labelled \textsc{Reasoning Correct} if it passes all checks up to small, non-critical deviations:
(1) correct identification of the global scaffold or molecular class (steroid, carbohydrate, amide, cyanine dye, thiazolium, etc.); if the scaffold is wrong from the outset (e.g.\ a sugar described as a steroid), the reasoning is immediately \textsc{Reasoning Flawed};
(2) recognition of the main functional groups (carboxylic acids, amides, phosphate diesters, sulfate esters, quaternary ammonium / thiazolium cations, glycosidic linkages, etc.) with roughly $80$--$90\%$ agreement with the gold molecule; severe confusions (e.g.\ phosphate vs.\ carboxylate, inventing a nitro group) are counted as flawed;
(3) internal logical consistency across steps (e.g.\ avoiding contradictions such as claiming an ``C$_{18}$ chain'' but assembling a C$_{16}$ fragment, or switching from ``monocyclic aromatic'' to ``two fused rings'');
and (4) the severity of chemical hallucinations.
We distinguish mild hallucinations (small, peripheral fragments that do not drive the construction) from severe ones (invented scaffolds or substituents that dominate the subsequent reasoning). In the former case the CoT may still be \textsc{Reasoning Correct} with minor noise; in the latter it is \textsc{Reasoning Flawed} with hallucination.

\begin{figure*}[t]
    \centering
    \includegraphics[width=0.48\textwidth]{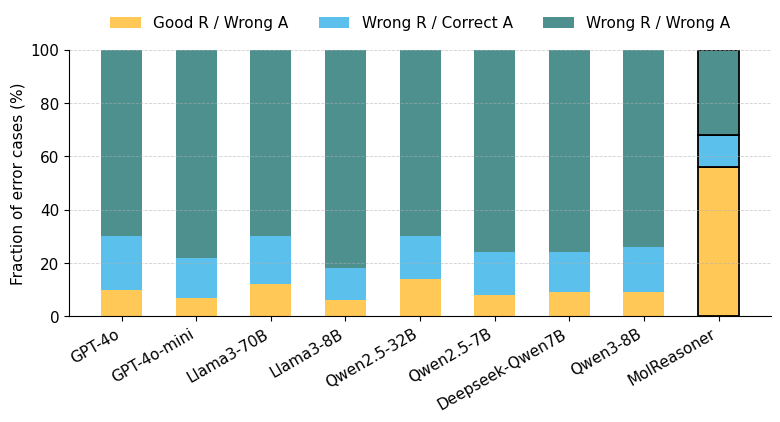}%
    \hfill
    \includegraphics[width=0.48\textwidth]{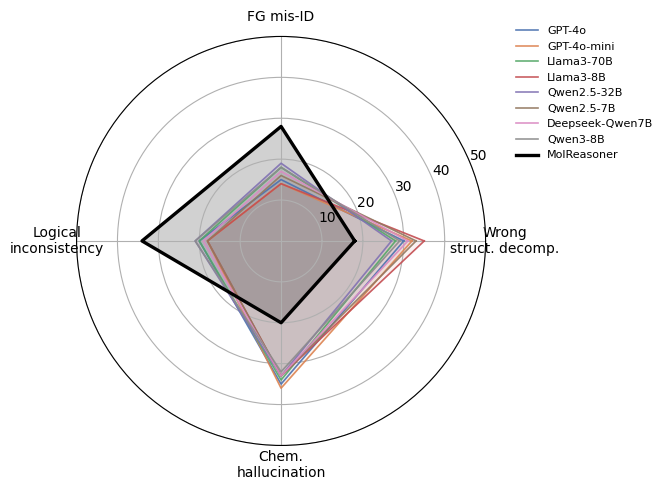}
    \caption{\textbf{Reasoning--answer coupling and error types on the \textsc{Molecule Captioning} (captioning) task.}
    Left: fraction of error cases that fall into ``good reasoning / wrong answer'', ``wrong reasoning / correct answer'', and ``wrong reasoning / wrong answer'' for each model.
    Right: breakdown of reasoning errors into four types: wrong structural decomposition, functional-group mis-identification, logical inconsistency, and chemical hallucination.
    MolReasoner reduces ``wrong reasoning / correct answer'' cases and shifts its errors toward more local, interpretable mismatches.}
    \label{fig:reasoning_answer_Molecule Captioning}
\end{figure*}

\begin{figure*}[t]
    \centering
    \includegraphics[width=0.48\textwidth]{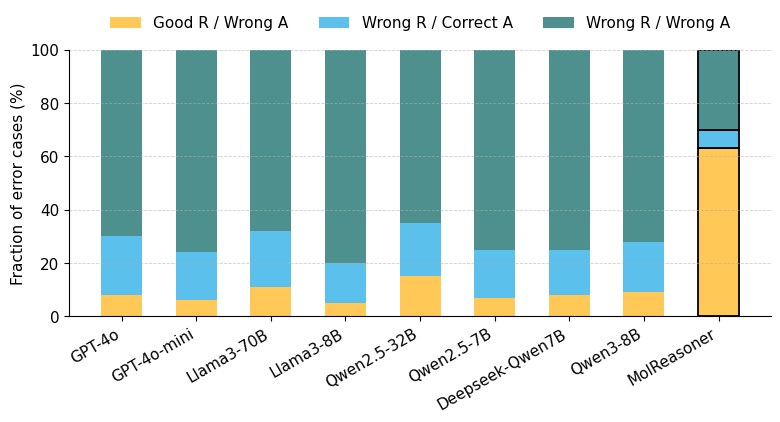}%
    \hfill
    \includegraphics[width=0.48\textwidth]{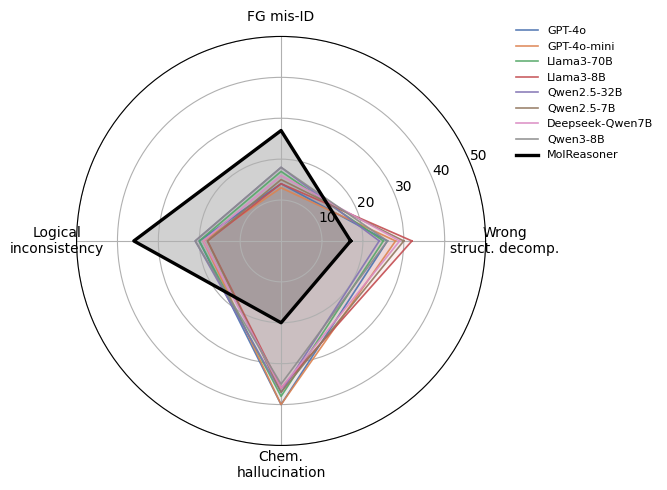}
    \caption{\textbf{Reasoning--answer coupling and error types on the \textsc{Text-based de novo Molecule Generation} (generation) task.}
    Analogous to Figure~\ref{fig:reasoning_answer_Molecule Captioning}, but for the generation task.
    Baselines are dominated by ``wrong reasoning / wrong answer'' and hallucination-type errors, whereas MolReasoner concentrates its failures in subtle functional-group mis-labelling and local assembly mistakes.}
    \label{fig:reasoning_answer_Text-based de novo Molecule Generation}
\end{figure*}

\textbf{Aggregate results.}
Figure~\ref{fig:reasoning_answer_Molecule Captioning} and Figure~\ref{fig:reasoning_answer_Text-based de novo Molecule Generation} show, for each model and task, the fraction of error cases that fall into the three categories:
``good reasoning / wrong answer'',
``wrong reasoning / correct answer'',
and ``wrong reasoning / wrong answer''.
Baseline models are dominated by the ``wrong reasoning / wrong answer'' regime, and also exhibit a substantial fraction of ``wrong reasoning / correct answer'' cases, where a correct molecule is obtained despite inconsistent or hallucinated CoT.
MolReasoner markedly reshapes this distribution: in both captioning and generation it has a much larger share of ``good reasoning / wrong answer'' cases and a much smaller share of ``wrong reasoning / correct answer'' cases than the baselines.

The radar plots in the same figures further decompose reasoning errors into four types:
(1) incorrect structural decomposition of the molecule;
(2) mis-identification of functional groups;
(3) logical inconsistency across steps; and
(4) chemical hallucination.
Baseline models show a large fraction of ``structural decomposition'' errors and hallucinations (inventing scaffolds or substituents not supported by the input), especially on the generation task. In contrast, MolReasoner strongly suppresses hallucination-type errors and shifts its mistakes toward more ``advanced'' modes: subtle functional-group mis-labelling and occasional logical inconsistencies in long CoT chains. Qualitatively, this matches our case studies: even when MolReasoner fails, it usually identifies the correct core scaffold and many of the correct functional groups, and the CoT is chemically coherent enough to diagnose where the assembly went wrong.

Taken together, these analyses support three conclusions.
First, we explicitly characterise the dominant failure modes (wrong decomposition, functional-group mis-identification, logical inconsistency, hallucination) and show that MolReasoner and generic LLMs fail in systematically different ways.
Second, there is a strong but not perfect coupling between reasoning quality and answer correctness: generic LLMs often ``guess right for the wrong reasons'', whereas MolReasoner shifts probability mass toward ``good reasoning / slightly wrong molecule''.
Third, by reducing catastrophic hallucinations and concentrating errors in interpretable, local mismatches, explicit molecular reasoning makes model failures easier to inspect and improves trustworthiness compared to black-box baselines.

\clearpage
\newpage
\section{Additional experimental results}\label{appx: exp_results}
In this section, we provide additional experimental results.

\subsection{Multidimensional Evaluation and
Qualitative Analysis}\label{appx: Multidimensional}
\begin{figure}[!t]
    \centering\hspace{0.4in}\includegraphics[width=\linewidth]{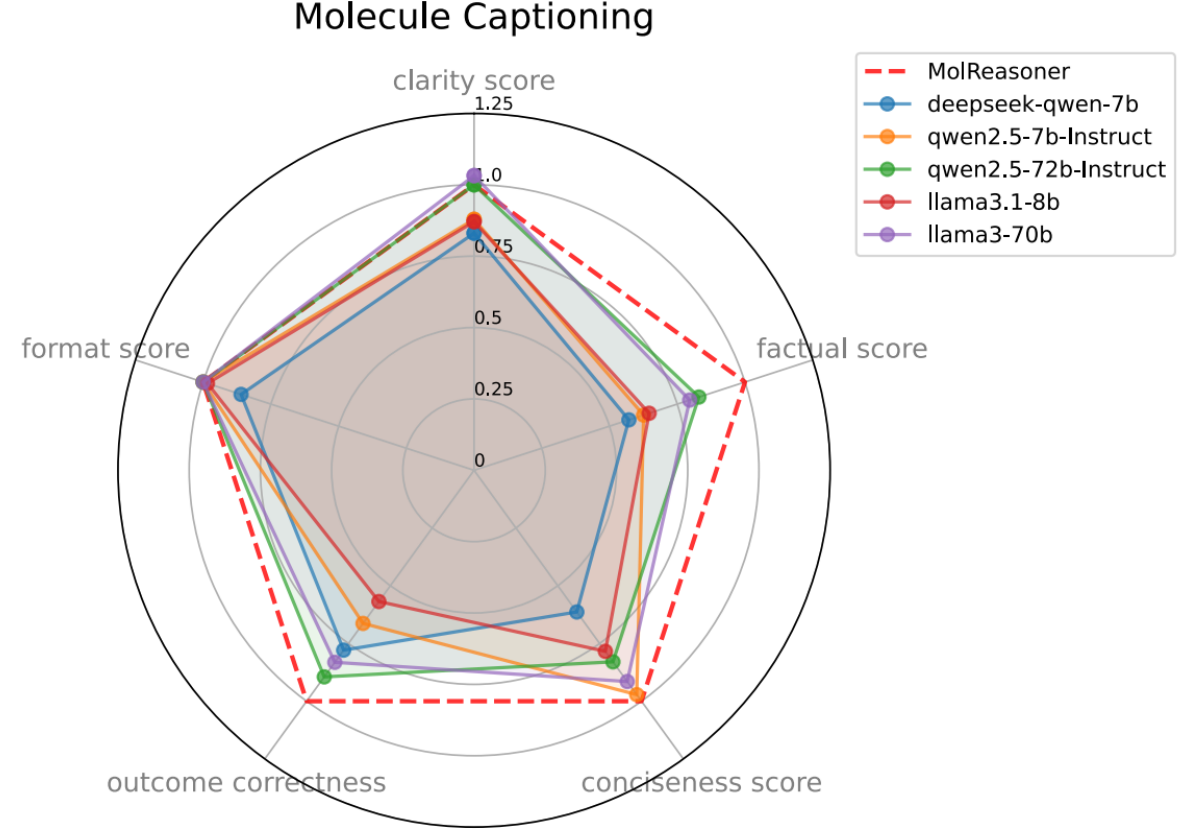} 
    \caption{\textcolor{black}
    {Performance of all models across five key evaluation metrics in the Molecule Captioning. To provide a more intuitive comparison, all scores are normalized by dividing them by the scores of MolReasoner.}}
    \label{fig: radar_mol2}
\end{figure}

Here we show the results of each model in Molecule Captioning in Figure \ref{fig: radar_mol2}.

\subsection{Individual Reward Ablation}\label{appx: Individual_appx}
\begin{figure}[!t]
\centering\includegraphics[width=\linewidth]{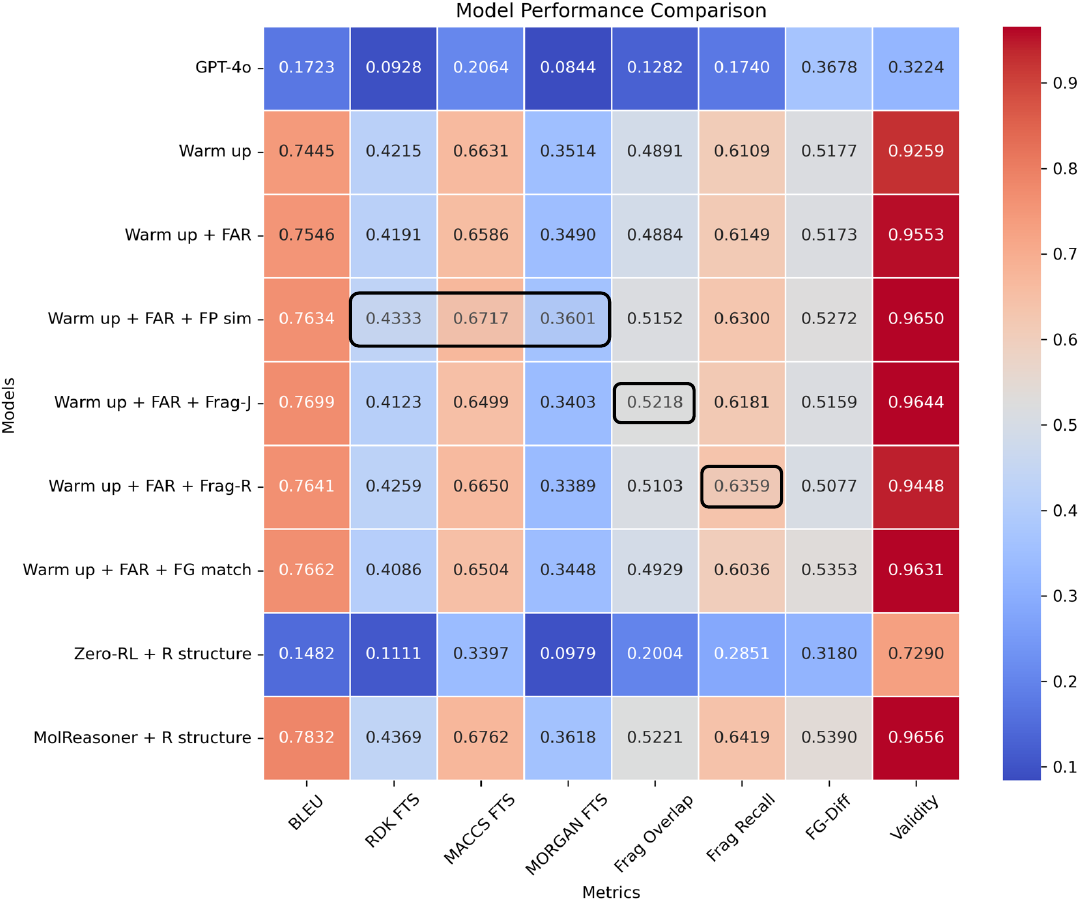}
    \caption{\textcolor{black}{Individual Reward Ablation For Text-based de novo Molecule generation.}}
    \label{fig: reward2}
\end{figure}

Here, we show the individual reward ablation results in Text-based de novo Molecule Generation in Figure \ref{fig: reward2}.

\subsection{Progressive Reward Composition Ablation}\label{appx: Progressive_appx}
Here we show the progressive reward composition ablation result for Text-based de novo Molecule Generation in Table \ref{tab:ablations_text_guided_molecule_generation}.

\begin{table*}[t]
  \centering
  \scalebox{0.75}{
  \setlength{\tabcolsep}{2.2pt}
    \begin{tabular}{@{}lccccccccccc@{}}
      \toprule
      \textbf{Method} & \textbf{BL.↑} & \textbf{Ex.↑} & \textbf{Le.↓} & \textbf{RDK.↑} & \textbf{MA.↑} & \textbf{MO.↑} & \textbf{Frag-J↑} & \textbf{Frag-R↑} & \textbf{FG-Match↑} & \textbf{Val.↑}  \\ \midrule
      \multicolumn{11}{c}{\textit{\textbf{Closed-Source Model}}} \\ \midrule
      GPT-4o & 0.1723 & 0.0062 & 50.2363 & 0.0928 & 0.2064  & 0.0844 & 0.1282  & 0.1740  & 0.3678 & 0.3224 \\ \midrule
      \multicolumn{11}{c}{\textit{\textbf{Ours}}} \\ \midrule
      \parbox[t]{4cm}{Warm-up} & 0.7444 & 0.0653 & 26.9821 & 0.4215  & 0.6630 & 0.3514 & 0.4890 & 0.6108 & 0.5177 & 0.9259 \\
      \parbox[t]{4cm}{+ FAR} & 0.7546 & 0.0734 & 27.2429 & 0.4191 & 0.6585 & 0.3490 & 0.4884 & 0.6149 & 0.5173 & 0.9552 \\
      \parbox[t]{4cm}{+ $\text{FP}_{\text{sim}}$} & 0.7636 & 0.0735 & 27.1251 & 0.4307 & 0.6616 & 0.3564 & 0.5023 & 0.6167 & 0.5180 & 0.9610 \\
      \parbox[t]{4cm}{+ $\text{FRAG}_{\text{sim}}$} & 0.7635 & 0.0741 & 27.1155 & 0.4293 & 0.6646 & 0.3587 & 0.5152 & 0.6346 & 0.5340 & 0.9612 \\
      \parbox[t]{4cm}{+ $\text{FG}_{\text{match}}$} & \underline{0.7682} & \underline{0.0743} & \underline{26.9617} & \underline{0.4346} & \underline{0.6753} & \underline{0.3603} & \underline{0.5169} & \underline{0.6374} & \underline{0.5385} & \underline{0.9632} \\
      \parbox[t]{4cm}{Zero-RL + $R_{\text{structural}}$} & 0.1482 & 0.0036 & 34.8775 & 0.1110 & 0.3396 & 0.0979 & 0.2004 & 0.2851 & 0.3180 & 0.7289 \\
      \rowcolor{mygray}
      \parbox[t]{4cm}{MolReasoner + $R_{\text{structural}}$} & \textbf{0.7832} & \textbf{0.0746} & \textbf{26.0237} & \textbf{0.4369} & \textbf{0.6762} & \textbf{0.3618} & \textbf{0.5221} & \textbf{0.6419} & \textbf{0.5390} & \textbf{0.9655} \\
      \bottomrule
    \end{tabular}
    }
  \caption{Progressive reward composition ablation study on different reward functions and the effect of warm-up stages for Text-based de novo Molecule Generation. "FAR" stands for Format Accuracy Reward, "$\text{FP}_{\text{sim}}$" refers to the fingerprint similarity combining RDK, MACCS, and MORGAN, "$\text{FRAG}_{\text{sim}}$" refers to the fragment similarity score combining Frag-J and Frag-R, "$\text{FG}_{\text{match}}$" refers to the functional group matching score, and "Zero-RL" indicates the model trained without warm-up. BL., Ex., Le., RDK., MA., MO., and Val. stand for BLEU, Exact, Levenshtein, RDK FTS, MACCS FTS, MORGAN FTS,  and Validity, respectively.}
  \label{tab:ablations_text_guided_molecule_generation}
\end{table*}

\subsection{Impact of Molecular Representation: SMILES vs. SELFIES}\label{appx: smiles_appx}
\begin{figure}[t]
\centering\includegraphics[width=\linewidth]{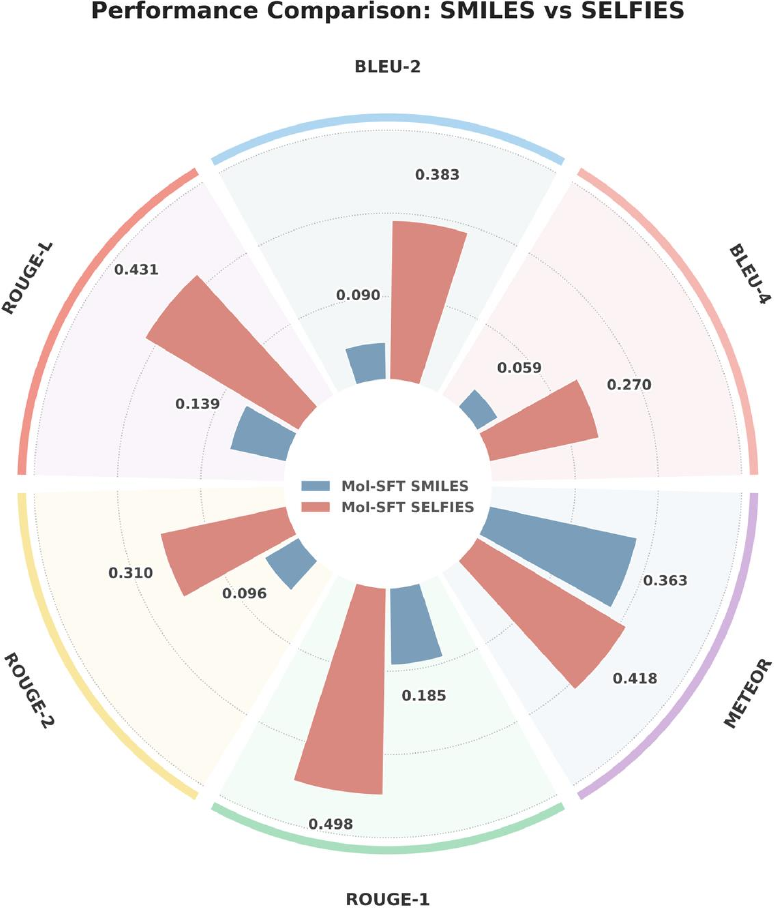} 
    \caption{\textcolor{black}
    {Impact of Molecular Representation.}}
    \label{fig: selfies}
\end{figure}

Here we show the comparison of molecular representation methods, SMILES and SELFIES, in Figure \ref{fig: selfies}.

\subsection{Out-of-Distribution Evaluation}\label{appx: ood_appx}
\begin{figure}[h]
    \centering\includegraphics[width=\linewidth]{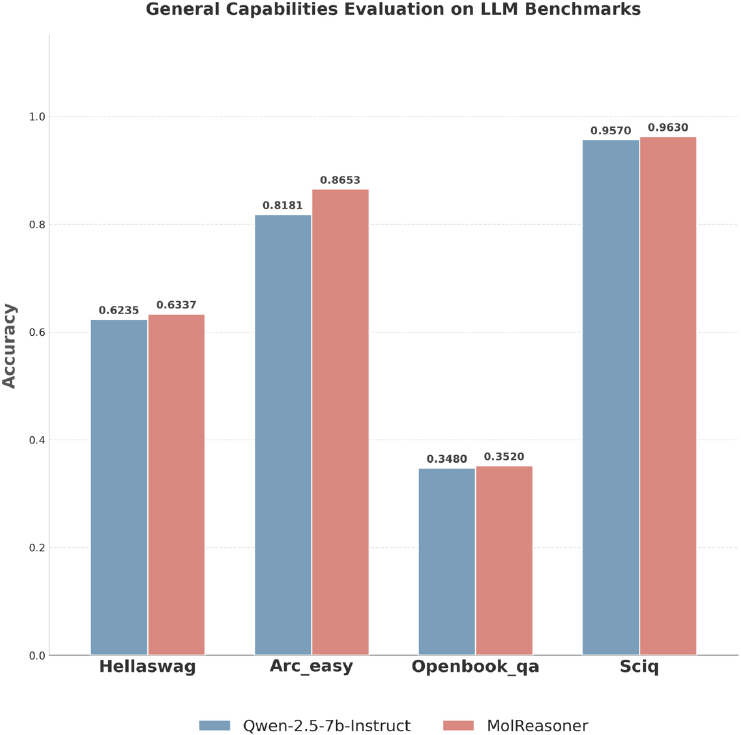} 
    \caption{\textcolor{black}
    {Out-of-Distribution Evaluation in General Domain Dataset.}}
    \label{fig: ood1}
    \vspace{-0.2in}
\end{figure}

\begin{figure*}[t]
    % \vspace{-0.2in}
    \centering\includegraphics[width=0.5\linewidth]{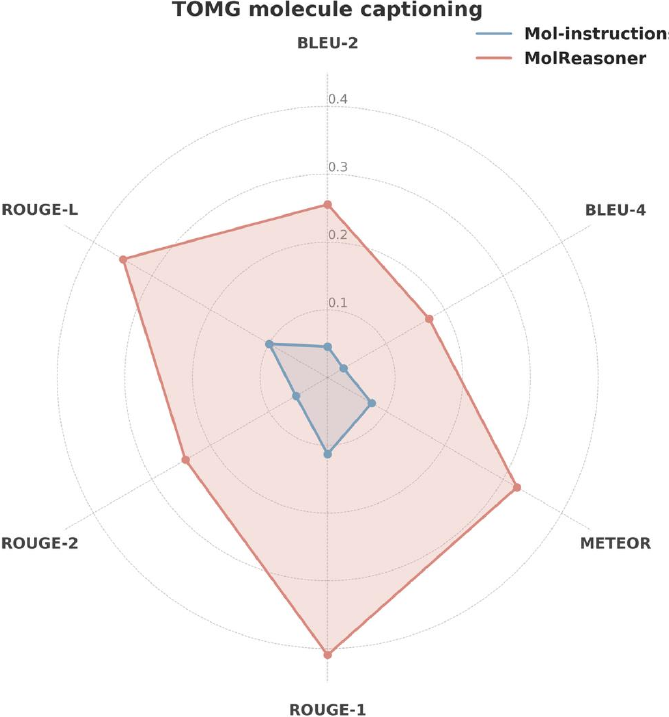} 
    \caption{\textcolor{black}
    {Out-of-Distribution Evaluation in Molecular Dataset.}}
    \label{fig: ood2}
    \vspace{-0.2in}
\end{figure*}

Here we show the comparison results of  molecular OOD 
dataset, in Figure \ref{fig: ood1} and \ref{fig: ood2}.

\end{document}